\documentclass[twoside]{article}

\usepackage[accepted]{aistats2025}
\usepackage[round]{natbib}
\renewcommand{\bibname}{References}
\renewcommand{\bibsection}{\subsubsection*{\bibname}}

\usepackage[utf8]{inputenc} 
\usepackage[T1]{fontenc}    
\usepackage{hyperref}       
\usepackage{url}            
\usepackage{booktabs}       
\usepackage{amsfonts}       
\usepackage{nicefrac}       
\usepackage{microtype}      
\usepackage{xcolor}         
\usepackage{amsmath}
\usepackage{hyperref}
\usepackage[capitalize,noabbrev]{cleveref}
\usepackage{amssymb}
\usepackage{mathtools}
\usepackage{mathrsfs}
\usepackage{amsthm}
\usepackage{dsfont}
\usepackage{xtab}
\usepackage{subcaption}
\usepackage{xfrac}
\usepackage{tikz}

\theoremstyle{plain}
\newtheorem{theorem}{Theorem}[section]

\newtheorem{lemma}[theorem]{Lemma}

\theoremstyle{definition}

\theoremstyle{remark}

\begin{document}

%

%

\twocolumn[

\aistatstitle{Legitimate ground-truth-free metrics for deep uncertainty classification scoring 
}

\aistatsauthor{ Arthur Pignet \And Chiara Regniez  \And John Klein }

\aistatsaddress{Owkin Inc. \And  Owkin Inc. \And Owkin Inc.} ]

\begin{abstract}
  Despite the increasing demand for safer machine learning practices, the use of Uncertainty Quantification (UQ) methods in production remains limited. This limitation is exacerbated by the challenge of validating UQ methods in absence of UQ ground truth.
    In classification tasks, when only a usual set of test data is at hand, several authors suggested different metrics that can be computed from such test points while assessing the quality of quantified uncertainties. 
    This paper investigates such metrics and proves that they are theoretically well-behaved and actually tied to some uncertainty ground truth which is easily interpretable in terms of model prediction trustworthiness ranking. 
    Equipped with those new results, and given the applicability of those metrics in the usual supervised paradigm, we argue that our contributions will help promoting a broader use of UQ in deep learning.
\end{abstract}

\section{INTRODUCTION}

Uncertainty quantification (UQ) is a research topic in machine learning that is gaining traction due to the increasing need for alignment with reliability requirements arising from safety critical application fields such as healthcare or autonomous driving. 
However, it is often overlooked and even seldom used in production. 
We argue that one of the main reasons behind this disparity is the lack of connections between uncertainty performance metrics and interpretable uncertainty "ground truth". 
Filling this gap would allow machine learning practitioners to better know to what extent the predicted uncertainty is actionable and this paper is an attempt to pave the way toward this goal.

To meet this ambition, a first impediment is that there is no universally agreed upon definition of uncertainty \citep{lahlou2021deup} which impairs the formalization of uncertainty quantification ground truth. 
This work aligns with \citet{kirchhof2023url} and regards uncertainty quantification as an estimation problem of the trustworthiness $s \left( \mathbf{x} \right) $ of a model's prediction for any input $\mathbf{x}\in \mathcal{X}$. 
The function $s$ is referred to as the (uncertainty) scoring function in the sequel.

Equipped with this vision, uncertainty quantification ground truth can be regarded as a function $s^*$ that is interpretable and objectively informs on the correctness of the model predictions. 
Let $\left( X, Y \right) $ denote an input-target pair of random variables governed by the data generative distribution $p_{X, Y}$. 
A realization of $\left( X, Y \right) $ is denoted by $\left( \mathbf{x}, y \right) \in \mathcal{X}\times \mathcal{Y} $. 
For convenience, we will use interchangeably $P \left( X=\mathbf{x}, Y=y \right) $ and $p_{X, Y}\left( \mathbf{x}, y \right) $.
We focus on binary classification tasks and $\mathcal{Y} = \left\{ 0;1 \right\}$. 

For classification problems, the minimizer of the posterior expected 0-1 loss is the Bayes classifier:
\begin{align}
    \hat{y}_{\textrm{bay}} \left( \mathbf{x} \right)  &= \underset{y \in \mathcal{Y}}{\arg\max}\: P \left(Y= y | X= \mathbf{x} \right), \forall \mathbf{x} \in \mathcal{X}.
\end{align}

If a training algorithm would succeed in learning the true conditional distribution of class labels given inputs, one would automatically gain access to an objective way to quantify how certain is the correctness of the model prediction $\hat{y}_{\textrm{bay}} \left( \mathbf{x} \right) $. Indeed, we have

\begin{align}
    \textrm{err}\left( \mathbf{x}; \hat{y}_{\textrm{bay}}, \boldsymbol\pi \right) &= P \left( Y\neq \hat{y}_{\textrm{bay}} | X= \mathbf{x} \right) \\ & = 1- \pi_{\hat{y}_{\textrm{bay}}}\left( \mathbf{x} \right) , \forall \mathbf{x} \in \mathcal{X},\label{eq:local_acc}
\end{align}

where $\boldsymbol\pi\left( \mathbf{x} \right)$ is such that $Y|X=\mathbf{x} \sim \text{Mult}\left( \boldsymbol\pi \left( \mathbf{x} \right)  \right)$.

The above scoring function is often referred to as the Bayes error and captures aleatoric uncertainty, i.e. the irreducible uncertainty carried by $p_{X, Y}$. As such, it is a legitimate ground truth $s^*$ to quantify the uncertainty of the Bayes classifier. Note that the UQ question is trivially solved in this ideal case since the Bayes classifier leverages the very probabilities $P \left(Y= y | X= \mathbf{x} \right)$, from which this ground truth is obtained. 

Deep learning leverages probabilistic classifiers $\hat{\boldsymbol\pi}$, where $\hat{\boldsymbol\pi} \left( \mathbf{x} \right) $ typically represents a vector of probabilities generated by a softmax layer, serving as an approximation of $\boldsymbol\pi\left( \mathbf{x} \right)$. 
For input $\mathbf{x} $, the predicted class label is $\hat{y} = \arg\max_{y \in \mathcal{Y}}\: \hat{\pi}_y\left( \mathbf{x} \right) $. 
In practice, due to epistemic uncertainty, $\hat{\boldsymbol\pi} $ deviates from $\boldsymbol\pi$. 
The gap between $\hat{\boldsymbol\pi}\left( \mathbf{x} \right)$ and the conditional distribution $p_{Y | \mathbf{x}}$ can be substantial depending on the amount of training data and modeling hypotheses. 
Based on this observation, two simple candidates for $s^*$ arise from equation~\eqref{eq:local_acc} in the context of UQ for a trained model $\hat{\boldsymbol\pi}$:
\begin{itemize}
    \item By substituting classifier $\hat{y}_{\textrm{bay}}$ with classifier $\hat{y}$, we obtain the probability of misclassification
    \begin{align}
        \phi \left( \mathbf{x} \right) &:=\textrm{err}\left( \mathbf{x}; \textcolor{blue}{\hat{y}} , \boldsymbol\pi \right)  \\ &= P \left( Y\neq \textcolor{blue}{\hat{y}} | X= \mathbf{x} \right) \\ &=  1 - \pi_{\textcolor{blue}{\hat{y}} }\left( \mathbf{x} \right) , \forall \mathbf{x} \in \mathcal{X}.\label{eq:misclassif_error}
    \end{align}
    \item By substituting the actual probabilities with softmax entries, we obtain an assessment of the commitment of the model to non-optimal decisions
    \begin{align}
        \varphi \left( \mathbf{x} \right) &:=\textrm{err}\left( \mathbf{x}; \hat{y}_{\textrm{bay}}, \textcolor{orange}{\hat{\boldsymbol\pi }} \right) \\ &= 1 - \textcolor{orange}{\hat{\pi}}_{\hat{y}_{\textrm{bay}}}\left( \mathbf{x} \right) , \forall \mathbf{x} \in \mathcal{X}.%
        \label{eq:p_gap}
    \end{align}    
\end{itemize}

Even endowed with definitions \eqref{eq:misclassif_error} and \eqref{eq:p_gap}, a major difficulty in UQ is the impossibility to evaluate those in real-world datasets or uncertainty ground truth in general \citep{kong2023tuto}. 
In the supervised learning paradigm, only $\left( \mathbf{x}, y \right)$ pairs are available, leaving the practitioner with no choice but to compute metrics using that test data. 
This paper focuses on the following angle in UQ evaluation: a purely metric based assessment of a UQ scoring function is valid if the metrics are maximized by one of the ground truth definitions \eqref{eq:misclassif_error} and \eqref{eq:p_gap}. 
We prove that one widely used metric in UQ benchmark is maximized by \eqref{eq:misclassif_error}. 
We also introduce a new metric that is is maximized by \eqref{eq:p_gap}. 

To sum up, our contributions are the following:
\begin{itemize}
    \item We provide theoretically grounded evidence that UQ-AUC - a metric used in most UQ benchmarks - is a legitimate way to assess the correctness of the quantified uncertainties even in absence of identified uncertain subsets of data points or distributional hypotheses.
    \item We shed light on important connections between UQ-AUC and other related prior arts. 
    \item We introduce a new metric, UQ-C-INDEX, which is maximized by a different scoring ground truth and offers an alternative standpoint for UQ evaluation.
    \item We prove that UQ-AUC and UQ-C-INDEX are actionable tools to identify subsets of data points for which the model reliability is higher.
    \item We provide an empirical evaluation of the suggested metrics on synthetic data with known scoring ground truth.
    \item We demonstrate the flexibility and minimalism of the setting in which those metrics can be used in production on real datasets. 
\end{itemize}
The investigated metrics can be generalized to the multi-class case through the usual one-versus-rest extensions. 

\subsection{Scope} 
The literature around UQ in deep learning is vast and the following paragraphs explain which part of this literature do not fall within the scope of the present paper.

\paragraph{This paper is not a UQ benchmark}
Since many UQ techniques were introduced (see \cref{ap:uq_overview} for an overview), several studies were conducted to benchmark them \citep{ovadia2019can,galil2023framework,galil2023can,kirchhof2023url}. 
Those benchmarks are insightful as they not only illustrate the limitations of UQ techniques but also offer guidance to more optimally leverage UQ techniques. 
We clarify that the present paper is not a benchmark. 
Our objective is to investigate uncertainty estimator metrics, some of which are actually used in the above cited benchmarks, from a theoretical standpoint.
By legitimizing a widely used metric and introducing a new one, our contributions reinforce the conclusions drawn in past benchmarks and provide insights for future ones.

\paragraph{This paper does not deal with OOD data}%
\label{sub:uq_use_cases}
There are two dominant use cases showcased in deep learning UQ articles: (i) ranking inputs based on classification error risk (ii) identifying out-of-domain (OOD) inputs. 
The former is a legitimate expectation from a well-behaved quantification, grading each model prediction likeliness based on patterns identified in inputs. 
The rationale behind the latter is that OOD data exhibit peculiar patterns compared to training data and have low likelihood under $p_{X,Y}$. 
Consequently, data uncertainty, i.e. epistemic uncertainty arising from limited training sample size, should be high in the vicinity of OOD data. 

Input ranking is the focus of this paper so as to fit to the classical ML setting where test data are iid with respect to the training data distribution $p_{X,Y}$. 
Indeed, the UQ metrics that we examine in this paper only require samples from the data generative distribution to be computed, which is always possible on real data by resorting to cross-validation. 
Conversely, the OOD detection use case requires extra OOD data that is not always at hand. 
However, if some are, elements of the proposed metric-based evaluation scheme remain applicable and this is discussed in \cref{ap:ood}. 

\section{SHEDDING THEORETICAL LIGHT ON METRICS FOR UQ EVALUATION}

The present section first provides arguments further legitimizing functions $\phi$ and $\varphi$ as relevant uncertainty scoring ground truths. 
It then examines a known scoring function metric from the literature and sheds light on its ties to ground truth $\phi$ in spite of requiring only access to test data points for computation. 
It also introduces a new metric enjoying similar properties, but this time being tied to ground truth $\varphi$. 
Note that while \citet{wimmer2023quantifying} examined the properties of scoring functions themselves, we examine the properties of metrics used to evaluate scoring functions.

The metrics analyzed in this section are meant to evaluate the ability of a scoring $s$ to rank inputs $\mathbf{x}$ from least to most likely to lead to a deviation of $\hat{y} \left( \mathbf{x} \right) $ from $\left\{ Y | X = \mathbf{x} \right\}$ or $\hat{y}_{\text{bay}}\left( \mathbf{x} \right)  $.

To investigate the goodness of ranking, it is useful to define, within the set $\mathcal{M}_X$  of $p_X$-measurable functions, the equivalence class $\mathcal{E}_s$ of scorings that almost surely rank inputs the same way as $s$: 
\begin{align}
    \mathcal{E}_s := &\left\{ s' \in \mathcal{M}_X \middle| P \left( \left\{ s \left( X \right) \geq  s \left( X' \right)\right\}\cap \left\{ s' \left( X \right)    \right. \right.  \right.   \nonumber \\  
    &\quad\quad\quad \left.\left. < s' \left( X' \right) \right\}  \right) = P \left( \left\{ s \left( X \right) <  s \left( X' \right)\right\}\cap \right. \nonumber \\ 
    &\quad\quad\quad\quad\quad\quad   \left.\left.\left\{ s' \left( X \right) \geq  s' \left( X' \right) \right\}  \right)=0 \right\}.
\end{align}

\subsection{Validity of $\phi$ and $\varphi$ to filter inputs based on uncertainty risks}
\label{sub:optimality_of_functions_}
Before unveiling how UQ metrics can assess the closeness between a scoring function $s$ and either ground truth $\phi$ or $\varphi$, let us first legitimize those ground truths in the context of input scoring based on predictive uncertainty. 
We propose to meet this objective by proving that sub-level sets of functions $\phi$ and $\varphi$ can gradually filter out inputs mapped to incorrect class labels (in some sense to be formalized below). 

In other words, we check if the investigated ground truth are actionable to define input space regions in which risk can be controlled. 

Let $\mathscr{C}_{\textrm{mce}} \left( \mathcal{E}_s \right) $ denote the set of values achieved by the \textbf{misclassification error} $\text{\textcolor{blue}{MCE}}_s \left( \beta \right)  = P \left( \textcolor{blue}{Y}  \neq \hat{Y} | s \left( X\right) \leq \beta   \right)$, $\forall\beta\in\mathbb{R}$ and $\mathscr{B}_{\textrm{mce}} \left( s \right) = \left\{ \min \left( \text{MCE}_s^{-1} \left( c \right)\right)  \middle|  c \in \mathscr{C}_{\textrm{mce}}\left( \mathcal{E}_s \right)  \right\}$ the set of minima of pre-images of $\text{MCE}_s$. 
The following result provides insights on the capability of sets to filter out inputs for which model predictions are unreliable. 

\begin{theorem}\label{thm:proof_nested_sets_phi}
 
$\forall s\in \mathcal{E}_{\phi}, \exists\psi_s$ a strictly increasing mapping from $\mathscr{C}_{\textrm{mce}}\left( \mathcal{E}_\phi \right) $ to $\mathscr{B}_{\textrm{mce}} \left( s \right) $  such that
\begin{align}
    P \left( \textcolor{blue}{Y} \neq \hat{Y} \middle|  s \left( X \right) \leq \psi_s \left( \gamma \right)   \right) = \gamma, \label{eq:nested_sets_phi}
\end{align} 
$\forall \gamma \in \mathscr{C}\left( \mathcal{E}_\phi \right) \subset \left( 0; 1-\text{ACC} \right].$
\end{theorem}

In the above theorem ACC stands for the accuracy of predictor $\hat{y}$. 
A proof of this result is given in \cref{ap:proof_nested_sets_phi}. 
A similar result for ground truth $\varphi$ follows. 
Let us introduce $\mathscr{C}_{\textrm{mbc}} \left( \mathcal{E}_s \right) $ the set of values achieved by the \textbf{misalignment with the Bayes classifier} $\text{\textcolor{orange}{MBC}}_s \left( \beta \right)  = P \left( \textcolor{orange}{\hat{Y}_{\textrm{bay}}} \neq \hat{Y} | s \left( X\right) \leq \beta   \right)$, $\forall\beta\in\mathbb{R}$ and $\mathscr{B}_{\textrm{mbc}} \left( s \right) = \left\{ \min \left(  \text{MBC}_s^{-1} \left( c \right) \right) \middle| \forall c \in \mathscr{C}_{\textrm{mbc}} \left( \mathcal{E}_s \right)  \right\}$.
\begin{theorem}\label{thm:proof_nested_sets_varphi}

$\forall s\in \mathcal{E}_{\varphi}, \exists\psi_s$ a strictly increasing mapping from $\mathscr{C}_{\textrm{mbc}} \left( \mathcal{E}_{\varphi}, \right) $ to $\mathscr{B}_{\textrm{mbc}} \left( s \right) $  such that
\begin{align}
    P \left( \textcolor{orange}{\hat{Y}_{\textrm{bay}}} \neq \hat{Y} \middle|  s \left( X \right) \leq \psi_s \left( \gamma \right) \right) = \gamma,  \label{eq:nested_sets_varphi}
\end{align}
$\forall \gamma \in \mathscr{C}_{\textrm{mbc}} \left( \mathcal{E}_\varphi \right)  \subset \left( 0; P \left( \hat{Y}_{\textrm{bay}} \neq \hat{Y} \right) \right].$
\end{theorem}
A proof of this result is given in \cref{ap:proof_nested_sets_varphi}. 

Simply put, these theorems state that ground truth scoring functions can span input space regions where the misclassification/misalignment probabilities can be confined to a prescribed value.
This is arguably one of the most desirable outcome of UQ. 
The input space regions at hand are the sub-level sets $\mathcal{L}_{\beta} \left( s \right)  := \left\{ \mathbf{x} \in \mathcal{X} \middle| s \left( \mathbf{x} \right) \leq \beta \right\}$ of a scoring function $s$.
To better understand the implication of \cref{thm:proof_nested_sets_phi}, one can also make connections with set-valued classifiers such as those issued by the conformal prediction framework~\citep{lei2015distribution,vovk2005algorithmic}. 
Indeed, let us introduce the following set-valued classifier 
\begin{align}
    \mathcal{C}_{\beta} \left(\mathbf{x} ; s \right) &= \begin{cases} \hat{y}\left( \mathbf{x} \right) & \text{if } \mathbf{x}\in \mathcal{L}_{\beta} \left( s \right)  \\ \mathcal{Y} & \text{otherwise} \end{cases}.
\end{align}

Then \cref{thm:proof_nested_sets_phi} says that, for any $s\in \mathcal{E}_{\phi}$, the set-valued classifier $\mathcal{C}_{\beta} \left(. ; s \right)$ can achieve any coverage $\gamma \in \mathscr{C}_{\textrm{mce}}\left( \mathcal{E}_s \right) $ by varying $\beta$, i.e. $P \left( Y \in \mathcal{C}_{\beta} \left(X ; s \right) \right) = \gamma$. 
This is also true for $\varphi$ but with a different flavor of coverage defined as $P \left( \hat{Y}_{\textrm{bay}} \in \mathcal{C}_{\beta} \left(X ; s \right) \right) $. 
Just like for conformalized predictors, coverage is not enough and one must check that, for different values $\beta$, sub-level sets $\mathcal{L}_{\beta} \left( s \right)$ capture a large amount of inputs which is equivalent to have set-valued predictors $\mathcal{C}_{\beta}\left( .; s \right) $ with small expected cardinalities.
This point will be addressed for each ground truth in the next subsections.

It can also be noted that \eqref{eq:nested_sets_phi} is reminiscent of confidence calibration. 
An important difference is that calibration is less flexible in the sense that it holds for a unique scoring function whereas \eqref{eq:nested_sets_phi} holds for infinite families of scorings. 
Moreover, the calibration property involves a conditioning on level sets of some the confidence score. 
Note that one could build a nested family of sets from a calibrated classifier to achieve \eqref{eq:nested_sets_phi} at the unavoidable cost of re-indexing.

\subsection{Checking that scoring functions match misclassification probabilities}

In practice, one has only access to a labeled (in-distribution) test dataset $\mathcal{D}_{\textrm{test}}$ of size $n_{\textrm{test}}$ sampled from the i.i.d. random variables $\left(Y^{(i)}, X^{(i)}\right)_{1\leq i \leq n_{\textrm{test}}}$. 
To evaluate the ability of a scoring function $s \left( \mathbf{x} \right)$ to align with misclassification probability $\phi \left( \mathbf{x} \right) $, it would be instrumental to leverage misclassification probabilities $\phi \left( \mathbf{x}^{(i)} \right)$ of the test points but those probabilities cannot be observed. 
One can thus wonder to what extent misclassification errors are a good proxy of $\phi \left( \mathbf{x}^{(i)} \right)$. 
Let us denote the misclassification function as follows:
\begin{align}
    \textrm{mis} \left( \mathbf{x}^{(i)} , y^{(i)}; \hat{y} \right)   = \mathds{1}_{ y^{(i)} \neq \hat{y}\left( \mathbf{x}^{(i)} \right)  } . \label{eq:crisp_misclassif}
\end{align}

One can then leverage misclassification errors to check if they match score $s$ by checking that,,
\begin{align}
    \textrm{mis}\left(\mathbf{x}^{(i)} , y^{(i)}; \hat{y} \right) &\leq \textrm{mis}\left( \mathbf{x}^{(j)} , y^{(j)}; \hat{y} \right)  \nonumber\\
    &\Leftrightarrow s \left( \mathbf{x}^{(i)}  \right) \leq s \left( \mathbf{x}^{(j)}  \right),\forall i \neq j.
\end{align}

Since misclassification errors are binary values, it is only worth checking the above equivalence for $\left( i, j \right) $ pairs such that $\textrm{mis}\left( \mathbf{x}^{(i)} , y^{(i)}; \hat{y} \right) \neq \textrm{mis}\left( \mathbf{x}^{(j)} , y^{(j)}; \hat{y} \right)$. 
In essence, this evaluation calculates the empirical probability that a correctly predicted instance has a lower uncertainty score than an incorrectly predicted one when both are chosen at random. 
ROC-AUC computes exactly empirical probabilities of this kind. 
We thus call this UQ metric the uncertainty quantification AUC and when $n_{\textrm{test}}\rightarrow \infty$, it writes:
\begin{align}
    \textrm{UQ-AUC}\left( s \right) &:= P \left( s \left( X^{(i)}  \right)  \leq s \left( X^{(j)}  \right) \middle|\textrm{mis}\left(X^{(i)}\right.\right., \nonumber\\
    &\quad\left.\left.  Y^{(i)} ; \hat{y} \right) \leq \textrm{mis}\left(   X^{(j)}, Y^{(j)} ; \hat{y} \right) \right).\label{eq:uq-auc}
\end{align}

This metric is precisely the one used in \citep{gal2016dropout,hendrycks2016baseline}. 
A first important theoretical result is that UQ-AUC is maximized for any score $s$ in the equivalence class of the misclassification probability. 

\begin{theorem}[Maximizing \textcolor{blue}{UQ-AUC}]\label{thm:uq_auc}
    A scoring function $s^*$ is a maximizer of UQ-AUC iff $s^* $ belongs to the equivalence class $ \mathcal{E}_{\textcolor{blue}{\phi}}$ of ground truth $\phi$.
\end{theorem}
A proof of this result, which builds upon~\citep[Proposition 1]{Clemencon2008ranking}, is given in \cref{ap:proof_thm_uq_auc}. 
What is particularly interesting with this result is that it holds for the actual misclassification probability ground truth without requiring access to misclassification probabilities $\phi \left( \mathbf{x}^{(i)} \right)$ for its computation.

As mentioned in the previous subsection, the legitimacy of $\phi$ as an uncertainty scoring ground truth also relies on its ability to produce sub-level sets $\left( \mathcal{L}_\beta \left( \phi \right) \right)_{\beta \in \left( 0;1 \right] } $ with a large probability to contain realizations of $X$. 
To be able to compare scoring functions in this regard, let us formalize this notion into the following preorder $\prec_{\text{mce}}$ : let $s$ and $s'$ be two scoring functions for which \eqref{eq:nested_sets_phi} holds, 
\begin{align}
    s \prec_{\text{mce}} s' \Leftrightarrow \mathcal{L}_{\psi_s \left( \gamma \right)  }\left( s \right)  \subseteq \mathcal{L}_{\psi_{s'} \left( \gamma \right)}\left( s' \right),\forall\gamma\nonumber\\
     \text{ and } \mu \left( \mathscr{R}_{ss'} \right)>0 
\end{align}
$\text{where } \mathscr{R}_{ss'} = \left\{ \gamma \in \mathscr{C} \middle|  P \left( X \in \mathcal{L}_{\psi_s \left( \gamma \right)} \left( s \right)  \setminus \mathcal{L}_{\psi_{s'} \left( \gamma \right)} \left( s' \right)\right.\right.$ $ \left.\left. | Y=\hat{Y} \right) > 0\right\}$ and $\mu$ is the Lebesgue measure. 
It can be proved that no other scoring $s$ for which the condition of \cref{thm:proof_nested_sets_phi} hold can uniformly dominate a member $s^*$ of $\mathcal{E}_\phi$. 

\begin{theorem}\label{thm:phi_optim}
  For any $s^*\in \mathcal{E}_{\textcolor{blue}{\phi}}$ and any other scoring $s$ for which \eqref{eq:nested_sets_phi} holds we have $ s^* \not\prec_{\text{\textcolor{blue}{mce}}} s $.
\end{theorem}
A proof of this result is given in \cref{ap:proof_phi_optim} which was the missing piece to fully establish the legitimacy of $\phi$ and as an uncertainty scoring ground truth.

\subsection{Checking that scoring functions catch misalignment with labels predicted by the Bayes classifier}
\label{sub:c-index}

In the following paragraphs, we introduce a new metric for scoring functions which is meant to catch inputs for which high softmax scores were assigned to non-optimal class labels, i.e. different from $\hat{y}_{\textrm{bay}}$.  
This new metric relies on a misclassification gap defined as
\begin{align}
    \Delta \left( \hat{\boldsymbol\pi} \left( \mathbf{x} \right) , y \right) :& = 1 - \hat{\pi}_y \left( \mathbf{x} \right)  .\label{eq:delta}%
\end{align}

By substituting function $\textrm{mis}$ with function $\Delta$ in \eqref{eq:uq-auc}, a new metric called UQ-C-index\footnote{In general, the concordance index also takes into account censorship but in the current use case there is no such censorship. This terminology is meant to emphasize that the metric relies on continuous functions.} is obtained:
\begin{align}
    \textrm{UQ-C-index}\left(s \right) := &P \left( s \left( X^{(i)}  \right)  \leq s \left( X^{(j)}  \right) \middle| \Delta \left( \hat{\boldsymbol\pi} \left(  X^{(i)} \right) \right.\right.\nonumber\\
    &\quad\quad\left.\left., Y^{(i)} \right) \leq \Delta \left( \hat{\boldsymbol\pi} \left(  X^{(j)} \right), Y^{(j)} \right) \right).\label{eq:uq-c-index}
\end{align}

Interestingly, this newly introduced metric is maximized by any member of the equivalence class of ground truth $\varphi$ as stated in the following theorem.
\begin{theorem}[Maximizing \textcolor{orange}{UQ-C-index}]\label{thm:uq_c_index} 
    Suppose $p_{\hat{\Pi}}=\hat{\boldsymbol\pi}\#p_X$ has no atom, where $\hat{\boldsymbol\pi}\#p_X$ is the push-forward of $p_X$ by $\hat{\boldsymbol\pi}$. 
    Then, a scoring function $s^*$ is a maximizer of UQ-C-index iff $s^* $ belong to the equivalence class $ \mathcal{E}_{\textcolor{orange}{\varphi}}$ of ground truth $\varphi$.
\end{theorem}
See \cref{ap:proof_uq_cindex_optimal_score} for a proof. 
This result shows that maximizing UQ-C-index is focused on checking the alignment between model probabilistic predictions with the predictions of the (optimal) Bayes classifier. 

Similarly as for the previous metric, let us introduce preorder $\prec_{\text{mbc}}$ : let $s$ and $s'$ be two scoring functions for which \eqref{eq:nested_sets_varphi} holds,
\begin{align}
    s \prec_{\text{mbc}} s' \Leftrightarrow \mathcal{L}_{\psi_s \left( \gamma \right)  }\left( s \right)  \subseteq \mathcal{L}_{\psi_{s'} \left( \gamma \right)}\left( s' \right),\forall \gamma \nonumber\\\text{ and } \mu \left( \mathscr{R}_{ss'} \right)>0. 
\end{align}

Observe that although the definition of $\mathscr{R}_{ss'}$ is the same as for pre-order $\prec_{\text{mce}}$, the two pre-orders are different because the functions $\psi_s$ and $\psi_{s'}$ involved in their respective definitions arise from property \eqref{eq:nested_sets_varphi} instead of \eqref{eq:nested_sets_phi}.
It can be proved that the sub-level sets spanned by members of $\mathcal{E}_{\varphi}$ exhibit a form of optimality stated in the following theorem. 

\begin{theorem}\label{thm:proof_nested_sets_c_index}
  Suppose $p_{S^*,\Delta} = \left( s^*, \Delta  \right)\# p_{X, Y}$ and $p_{S,\Delta} = \left( s, \Delta  \right)\# p_{X, Y}$ have no atom. 
  For any $s^*\in \mathcal{E}_{\textcolor{orange}{\varphi}}$ and any other scoring $s$ for which \eqref{eq:nested_sets_varphi} holds we have $s^* \not\prec_{\text{\textcolor{orange}{mbc}}} s $.
\end{theorem}

A proof of theorem \cref{thm:proof_nested_sets_c_index} can be found in \cref{ap:proof_nested_sets_c_index}.
This theorem was the missing piece to fully establish the legitimacy of $\varphi$ and as an uncertainty scoring ground truth alternative.

\begin{figure*}[!h]
    \centering
    \begin{subfigure}[b]{0.24\textwidth}
      \includegraphics[width=\textwidth]{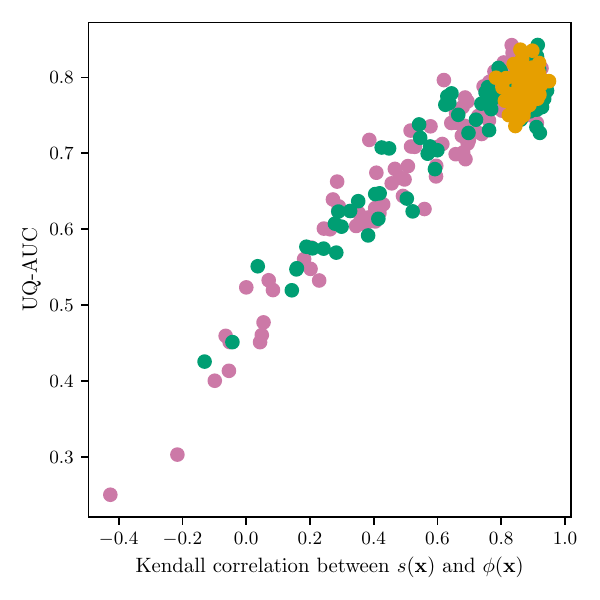}
      \caption{Synthetic\label{fig:aucuq-bayes-error-synth}}  
   \end{subfigure}
  \begin{subfigure}[b]{0.24\textwidth}
      \includegraphics[width=\textwidth]{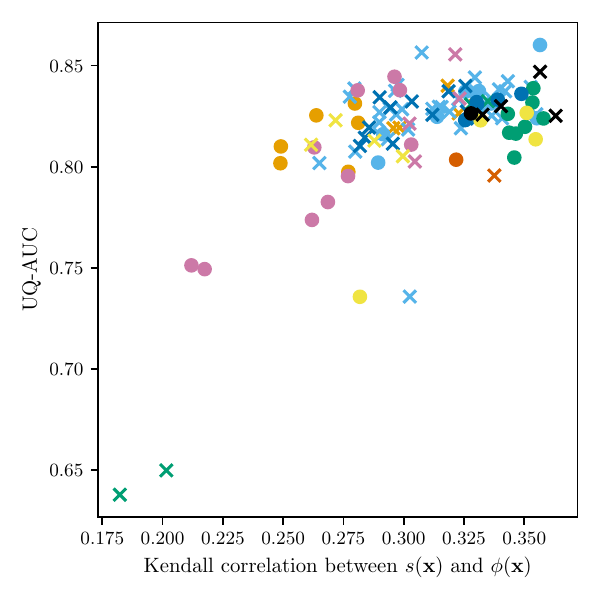}
       \caption{CIFAR10-H\label{fig:aucuq-bayes-error-cifar10h}}
      \end{subfigure}
    \begin{subfigure}[b]{0.24\textwidth}
       \includegraphics[width=\textwidth]{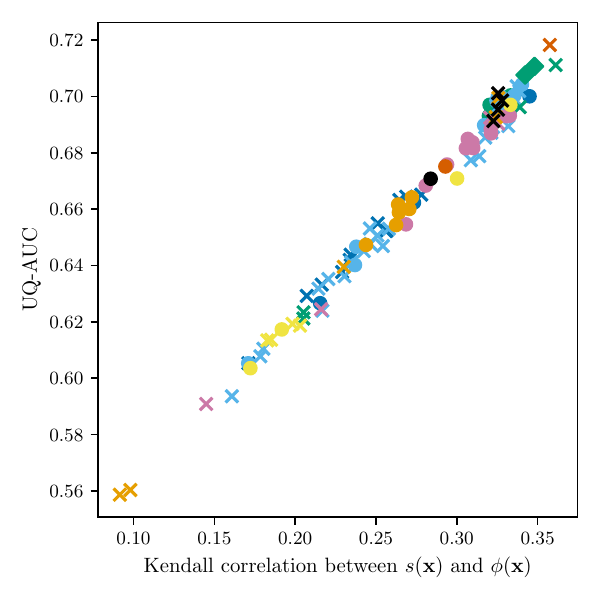}
       \caption{R-ImageNet\label{fig:aucuq-bayes-error-imagenet}} 
    \end{subfigure}
    \begin{subfigure}[b]{0.24\textwidth}
      \includegraphics[width=\textwidth]{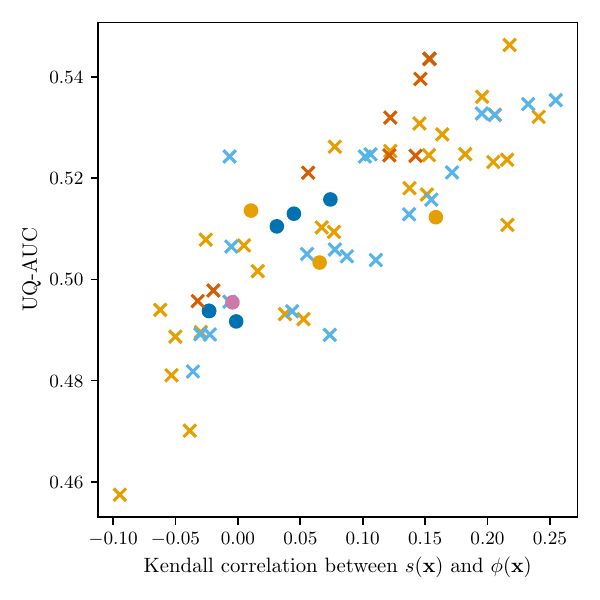}
      \caption{OASST2\label{fig:aucuq-bayes-error-oasst2}}
    \end{subfigure}
  
    \begin{subfigure}[b]{0.24\textwidth}
      \includegraphics[width=\textwidth]{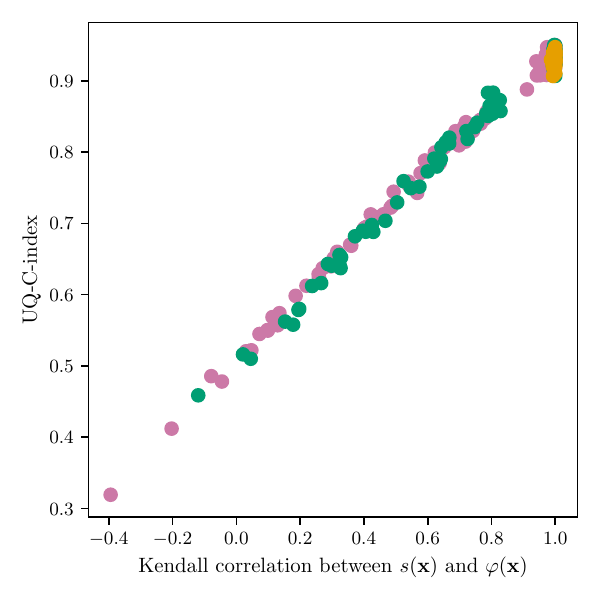}
      \caption{Synthetic\label{fig:cuq-misalign-bayes-label-synth}}
    \end{subfigure}
    \begin{subfigure}[b]{0.24\textwidth}
       \includegraphics[width=\textwidth]{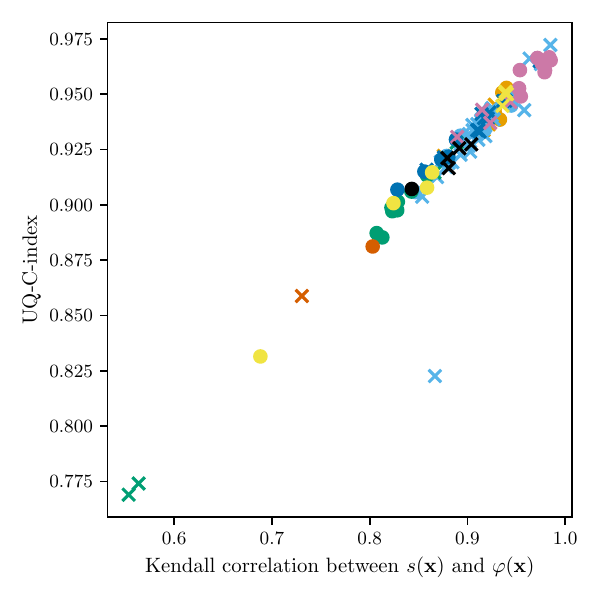}
       \caption{CIFAR10-H\label{fig:cuq-misalign-bayes-label-cifar10h}}
    \end{subfigure}
    \begin{subfigure}[b]{0.24\textwidth}
       \includegraphics[width=\textwidth]{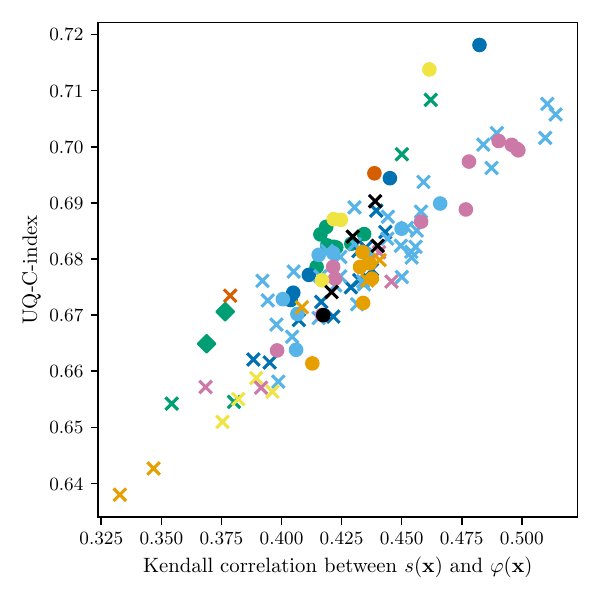}
       \caption{R-ImageNet\label{fig:cuq-misalign-bayes-label-imagenet}}
    \end{subfigure}    
    \begin{subfigure}[b]{0.24\textwidth}
       \includegraphics[width=\textwidth]{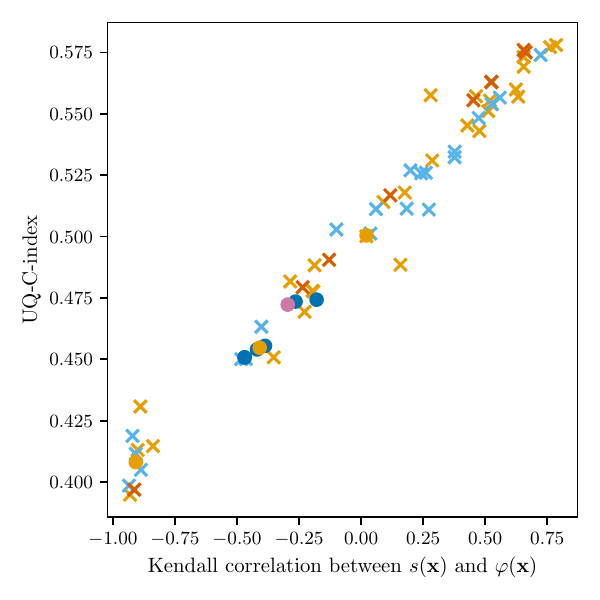}
       \caption{OASST2\label{fig:cuq-misalign-bayes-label-oasst2}}
    \end{subfigure}

    \caption{Synthetic dataset and real-world datasets with human annotations. Uncertainty ground truth-free metric UQ-AUC (Top row, resp. UQ-C-index in the bottom row) versus ground truth dependent Kendall correlation between $s(\mathbf{x})$ and $\phi(\mathbf{x})$, (resp. between $s(\mathbf{x})$ and $\varphi(\mathbf{x})$) for several UQ algorithms and hyperparameter configurations. All metrics are computed on a held-out test set. Note that Kendall correlations are intractable in practice, contrary to the UQ-AUC or UQ-C-index.\label{fig:metrics-human-annotations-real-world}. 
    \newline Models used for Synth: 
    Entropy \textcolor[HTML]{E69F00}{\textbullet}, 
    Deep Ensemble \textcolor[HTML]{009E73}{\textbullet}, 
    Monte-Carlo Dropout \textcolor[HTML]{CC79A7}{\textbullet}, 
    \newline Models used for CIFAR10 and R-ImageNet:  
    GoogleNet \textcolor[HTML]{D55E00}{\textbullet}, 
    AlexNet \textcolor[HTML]{D55E00}{$\times$}, 
    ResNet \textcolor[HTML]{56B4E9}{\textbullet}, 
    RegNet \textcolor[HTML]{56B4E9}{$\times$}, 
    Swin \textcolor[HTML]{E69F00}{\textbullet}, 
    ResNeXt \textcolor[HTML]{E69F00}{$\times$}, 
    MobileNet \textcolor[HTML]{0072B2}{\textbullet}, 
    EfficientNet \textcolor[HTML]{0072B2}{$\times$}, 
    ViT \textcolor[HTML]{CC79A7}{\textbullet}, 
    Wide ResNet \textcolor[HTML]{CC79A7}{$\times$}, 
    MnasNet \textcolor[HTML]{F0E442}{\textbullet}, 
    ConvNeXt \textcolor[HTML]{F0E442}{$\times$}, 
    Inception \textcolor[HTML]{000000}{\textbullet}, 
    DenseNet \textcolor[HTML]{000000}{$\times$}, 
    VGG \textcolor[HTML]{009E73}{\textbullet}, 
    ShuffleNet \textcolor[HTML]{009E73}{$\times$},
    \newline Models used for OASST2:
    BART \textcolor[HTML]{D55E00}{$\times$},
    DistilBERT \textcolor[HTML]{D55E00}{\textbullet},
    DeBERTa \textcolor[HTML]{000000}{$\times$},
    RoBERTa \textcolor[HTML]{56B4E9}{\textbullet},
    BERT \textcolor[HTML]{E69F00}{$\times$},
    XLM \textcolor[HTML]{E69F00}{\textbullet},
    MiniLM \textcolor[HTML]{0072B2}{$\times$},
    E5 \textcolor[HTML]{0072B2}{\textbullet},
    mContriever \textcolor[HTML]{CC79A7}{\textbullet},
    Other \textcolor[HTML]{56B4E9}{$\times$}},
  
  \end{figure*}
  \begin{table*}[!t]
    \caption{Pearson correlations between UQ metrics and ground-truth dependent Kendall coefficients $\kappa$. Each correlation value is followed by lower and upper 95\% confidence intervals, obtained using Fisher's z-transformation.
    Significantly better correlation is shown in bold letters.\label{tab:pearson}}%
    \label{sample-table}
    \begin{center}
    \begin{tabular}{llll}
    \bf Dataset & \multicolumn{1}{c}{\bf UQ metric}  &\multicolumn{1}{c}{\bf Correl. with $\kappa \left( s; \phi \right) $} & \multicolumn{1}{c}{\bf Correl. with $\kappa \left( s; \varphi\right) $}
    \\ \hline \\
    Synth. &UQ-AUC$(s)$         & 94.56\% (93.36;95.55) & 93.44\% (92.00;94.3) \\
           &UQ-C-index$(s)$     & \textbf{98.81}\% (98.54;99.03) & \textbf{99.62}\% (99.53;99.69) \\
    \hline\\
    CIFAR10-H &UQ-AUC$(s)$         & \textbf{65.18}\% (53.99;74.75) & 54.01\% (39.33;66.00) \\
              &UQ-C-index$(s)$     & -0.04\% (-22.41;14.83) & \textbf{95.61}\% (93.67;96.97) \\
    \hline\\
    R-ImageNet &UQ-AUC$(s)$         & \textbf{99.56}\% (99.36;99.69)  & 39.42\% 23.47;53.75)  \\
               &UQ-C-index$(s)$     & 40.20\% (23.56;54.56) & \textbf{88.07}\%  (83.17;91.63)  \\
    \hline\\
    OASST2 &UQ-AUC$(s)$         & 	\textbf{82.50}\% (73.10; 88.82) & 71.69\% (57.84; 81.53)  \\
    &UQ-C-index$(s)$     & 74.01\%  (61.04; 83.12) & \textbf{98.88}\% (98.19; 99.31) \\
  
  \end{tabular}
  \end{center}
  \end{table*}
  
\section{RELATED WORK}

If the role of an uncertainty scoring function is to provide insights regarding a model's trustworthiness, how can one be sure that scoring functions themselves are trustworthy?
ML practitioners would typically investigate this question on a held-out dataset as part of an evaluation phase. 
Unfortunately, assessing the quality of quantified uncertainties appears to be more challenging than other ML predictions as there is no genuine ground truth in the form of \eqref{eq:misclassif_error} or \eqref{eq:p_gap} which is mainly why no consensus on UQ metrics has emerged in the literature~\citep{ovadia2019can}. 
This performance evaluation challenge was addressed in prior arts (sometimes depending on the sought use case: input ranking or OOD detection). The following paragraphs provide an overview of these related works.

One line of UQ validation leverages a companion task that aims at leveraging meaningful signals in an uncertainty score. 
One such task is uncertainty sampling for active learning \citep{gal2017deep,nguyen2022measure}. 
If a UQ method correctly selects data points with high predictive uncertainty, revealing their labels should provide a strong supervisory signal and allow to maximize accuracy increase. 
At each active learning round, \citet{gal2017deep} select the input that maximizes an uncertainty scoring function, add it to the training set and re-train the model.  
When using the total entropy,  mutual information, or variation ratio function, the test accurracy increases at a pace significantly faster compared to random selection of the next input to insert in the training set. 
\citet{nguyen2022measure} conducted a broader range of similar experiments and show that, depending on the model class, uncertainty scoring functions capturing epistemic uncertainty rather than aleatoric uncertainty can be far better acquistion functions. 
In a similar spirit, UQ scoring can be used as feedback for the initial task. \citet{mortier2023calibration} build weighted ensembles where the weights of the ensemble members are computed from the scoring issued by the UQ technique. 
If a UQ technique is well behaved, its corresponding ensemble should promote reliable classifiers and achieve better classification performances. 
However, evaluation in this setting requires the uncertainty scores to be well calibrated. In contrast, this paper addresses the evaluation of scores regardless of their calibration performances. 
Reinforcement learning can also typically benefit from feedback and is used for validation in \citep{gal2016dropout}.
Specifically, in situations where the Q-value function is approximated by a neural network, predictive uncertainty can be leveraged to perform Thomson sampling in lieu of an $\epsilon$-greedy search. 
The higher the expected cumulative reward, the better the uncertainty estimates. 

A rigorous way to check if quantified uncertainties align with errors is to rely on controlled experiments that typically involve a physical system \citep{caldeira2020deeply}. Depending on the system in question, sources of aleatoric and/or epistemic uncertainty can be varied which will impact the learnt model emulating the physical system. 
In a sensitivity analysis spirit, one may check that a scoring function varies monotonically with each source and compute a correlation metric. 
However this objective evaluation remains grounded with a single regression task and does not reach the genericity objective to evaluate quantified uncertainties on arbitrary data and classification tasks.

In spite of their appeal, the above cited elements of the literature do no possess the sought level of genericity allowing for uncertainty estimator evaluation solely based on $\left( \mathbf{x}, y \right) $ pairs. 
Indeed all these methods are indirect measures of uncertainty scores, strongly dependent on the chosen backbone algorithm. 
In the next paragraph, we present metric based evaluation prior arts that comply with this requirement.

A popular way to check the legitimacy of predicted probabilities is to resort to calibration metrics \citep{nemani2023uncertainty,wimmer2023quantifying,jospin2022hands,henne2020benchmarking,ovadia2019can}. 
By definition, obtaining calibrated probabilities is a much stronger requirement than obtaining a score that correctly ranks unreliable predictions. 
Indeed, suppose one wishes to rank unreliable predictions thanks to calibrated softmax probabilities, then any composition of those probability with an increasing function will rank inputs in the same way. 
This invariance property of the investigated metrics compared to the expected calibration error is further highlighted in~\cref{sec:metrics_assessing_ranking_performance_versus_calibration_performance}. 

When assessing whether quantified uncertainties align with classification errors, many authors converged to a similar setting: leverage a scoring function and evaluate the latter as if it was meant to solve another binary classification task in which inputs are mapped to either $\left\{ Y=\hat{y}|\mathbf{x}\right\}$ or $\left\{ Y\neq\hat{y}|\mathbf{x}\right\}$ \citep{henne2020benchmarking,gal2016dropout,hendrycks2016baseline}. 
Given a held-out test dataset it is possible to construct those misclassification labels. 
Finally, ROC-AUC can readily be used, thereby yielding the UQ-AUC metric formally defined in the previous section.
Similarly, in the OOD detection use case, provided that some OOD inputs can be obtained, ROC-AUC of a UQ scoring with respect to in-distribution versus OOD labels can be computed \citep{gal2016dropout,hendrycks2016baseline,lakshminarayanan2017simple}

Finally, the closest elements of the literature compared to this paper are \citep{geifman2018bias} and \citep{henne2020benchmarking}, each of them introducing a UQ metric. The metric proposed in \citep{geifman2018bias} uses a threshold on scoring functions to filter out low score data and build a subset of higher score data. For each such subset, one can compute its size fraction (compared to the entire subset) and an accuracy. By browsing several threshold values, a curve is built and the metric value is the AUC of this curve. 
In the sequel, we not only prove that, for any fixed predictor $\hat{y}$, this metric is in bijection with UQ-AUC (thereby essentially grasping the same information as the latter) but also that it benefits from the same results as UQ-AUC. 
A similar remark holds for the work of \citet{henne2020benchmarking} who compare UQ techniques based on a curve whose AUC is  proportional to the UQ-AUC.

\subsection{Metric comparison synthesis} 
\label{sub:metric_comparison}

In the following paragraphs, the main differences between metrics investigated in the previous subsections and other closely related works are highlighted. 
We denote the metric from \citep{geifman2018bias} $\textrm{G-AUC}$ and the one arising from the work of \citep{henne2020benchmarking} $\textrm{H-AUC}$
Let us first formally establish the bonds of these AUCs with UQ-AUC.
\begin{lemma}\label{lem:aurc}
    For any scoring function $s$ and fixed predictor $\hat{y}$ with accuracy $\textrm{ACC}$, we have
    \begin{align}
        \textrm{G-AUC}\left( s \right)  = &\left( 1 - \textrm{ACC}  \right)^2 + 2 \left( 1 - \right. \label{eq:aurc}\nonumber\\
        &\quad \left.\textrm{UQ-AUC} \left( s \right)   \right)\left( 1 - \textrm{ACC}  \right) \textrm{ACC} , \\
        \textrm{H-AUC}\left( s \right)  = &\textrm{UQ-AUC} \left( s \right)  \left( 1 - \textrm{ACC}  \right) \textrm{ACC}.  \label{eq:h-auc}
    \end{align}    
\end{lemma}

The above lemma clarifies important connections between prior arts. 
It is clear that UQ benchmarks relying on either UQ-AUC, G-AUC or H-AUC will lead to identical conclusions. 
This also confirms the novelty of metric UQ-C-index compared to prior arts. 
It can be argued that the influence of the accuracy of $\hat{y}$ in the definitions of the G-AUC and H-AUC metrics make them harder to interpret than UQ-AUC. 
For example, the maximal value of those two metrics also depends on ACC and thus cannot be used as universal reference. 

Another important point to highlight are the different shades of scoring performance that UQ-AUC and UQ-C-index promote respectively. 
Those are in fact rooted in the definitions of the uncertainty ground truth to which each metric is tied and the different validity properties they enjoy (\cref{thm:proof_nested_sets_phi} and \cref{thm:proof_nested_sets_varphi} respectively). 
While UQ-AUC promotes scoring functions allowing to secure input space regions where the trained model issues correct class predictions, UQ-C-index promotes scoring functions allowing to secure input space regions where the trained model issues optimal class predictions. 
The difference between the two is rather subtle but it can be argued that UQ-C-index has the advantage to reward scoring functions that rank highly inputs that the model learnt to map to the most probable class labels and filters out correct labeling arising from lucky outcomes where $Y\neq \hat{y}_{\textrm{bay}}$ while having $Y = \hat{y}$. 
On a same note, when the aleatoric uncertainty vanishes, then properties \cref{thm:proof_nested_sets_phi} and \cref{thm:proof_nested_sets_varphi} become equivalent and the two metrics will promote similar scoring functions.

\section{EXPERIMENTS} 
\label{sec:experiments}

This section is dedicated to experimentally illustrating the soundness of the investigated UQ metrics\footnote{All code is available in the following repository: \url{https://github.com/owkin/legitimate-uq-metrics}.}.
In this setting, we emphasize that the proposed metrics are ground-truth free and can be used to evaluate UQ methods solely based on $\left( \mathbf{x}, y \right) $ pairs.
However for the sake of comparison, we need a ground-truth dependent metric, namely the Kendall correlation $\kappa \left( s; \textrm{gt} \right) $ between a scoring $s$ and ground truth $ \textrm{gt} \in \left\{ \phi, \varphi  \right\} $. 
We conducted our experiment first on synthetic data, where the uncertainty ground-truth is tractable by design. 
Then we scale our experiment on three real-world datasets, CIFAR10, R-ImageNet and OASST2, and leverage extra human annotations as a proxy for UQ ground-truth.
More details on the datasets and the models used are provided in \cref{app:exp-detail}.  
Extra experiments notably illustrating some of our theoritical findings are presented in \cref{app:exp-detail}.

In the supervised learning setting, the ground truths $\phi$ and $\varphi $ cannot be obtained from $\left( \mathbf{x}, y \right) $ pairs which is why this kind of evaluation can only be carried out (with maximal reliability) using synthetic data where the data generating distribution is known.
To obtain closed forms for $\phi  $ and $\varphi $, we design a toy binary classification dataset which is a mixture of two overlapping Gaussian distributions in input space, one for each label. The dataset is described at length in \cref{app:exp-detail}. 

We trained several UQ aware models, and for each UQ method, we vary the hyper-parameters and evaluate the performances of the obtained scoring functions using UQ-AUC, UQ-C-index and the Kendall correlation between the scoring function and either $\phi$ or $\varphi$. 
Scatter plots are reported in~\cref{fig:aucuq-bayes-error-synth,fig:aucuq-bayes-error-cifar10h}. 
In those plots, each dot corresponds to one function $s \left( \mathbf{x} \right) $. 
An excellent match with the ground-truth dependent coefficients can be observed which confirms that both the UQ-AUC and UQ-C-index are well-behaved metrics to evaluate the quality of the uncertainty scoring functions. 
This is confirmed by \cref{tab:pearson} which shows Pearson correlations between the two metrics with the Kendall coefficients depending on any of the ground truth. 
It is remarkable that very good correlations are obtained for each metric even with the ground truth it is not tied to and even if aleatoric uncertainty is significant in the data. 

An important message from those experiments is that the asymptotical qualities of the UQ-AUC and UQ-C-index metrics established by theoretical results from the previous section empirically propagate to the finite sample case.
Besides, \cref{fig:aucuq-bayes-error-synth,fig:aucuq-bayes-error-cifar10h} highlight that the choice of hyper-parameters (either of the model, training procedure or UQ method) is crucial to obtain meaningful scores. Indeed, several runs exhibit a negative Kendall correlation. 
The UQ-AUC and UQ-C-index experimentally have the desirable property that for negative Kendall correlation, their values are below 0.5 most of the time. 

For the experiment on real-world image datasets, we leverage 116 models trained on ImageNet. 
For CIFAR10-H, we fine-tuned those models on the ultimate layer for 10 epochs.  
Following the work of \citet{peterson2019human}, we leverage extra human annotations as a proxy for uncertainty ground truth, and evaluate the correlation between the scoring functions and the aforementioned human uncertainty. 
This is meant to check if the investigated metrics, without having access to ground truth, are still aligned with human judgments. 
We thus compare the investigated metrics UQ-AUC and UQ-C-index to Kendall correlations (between $s(\mathbf{x})$ and ground truths computed from $P(Y_{\text{human}}|\mathbf{x})$).
For the natural language dataset OASST2, 70 pre-trained models were leveraged. 
Several human annotations per text are also available to assess the text content as positive or negative thus enabling to build an uncertainty ground truth. 
Additional information on the real-world datasets and the leveraged models are provided in \cref{app:exp-detail}.

\cref{fig:aucuq-bayes-error-imagenet} to \cref{fig:cuq-misalign-bayes-label-imagenet} show the correlation of the investigated metrics with the uncertainty ground-truths they are tied to. 
Similarly as for synthetic data, high levels of correlations are observed between the Kendall coefficients of a ground truth and the metric maximized by this ground truth.
However, looking at \cref{tab:pearson} significantly lower correlations are observed when  Kendall coefficients of a ground truth are compared to the metric not maximized by them. 
This shows that the two metrics do not catch the same signals within scoring functions and offer different standpoints as to the performances of the those functions. 
We believe that this contrast is rooted in the very definitions of the ground truths, one seeking rewarding correct predictions, the other one rewarding correct decisions.

\section{CONCLUSION}%
\label{sec:conclusion}
One of the most popular outcomes of deep uncertainty methods are scoring functions which are meant to rank inputs based on their likeliness to be mapped to correct class labels by a trained model. 
A legitimate expectation of some end-user of such scoring functions is to be able to check if the issued ranking are effectively allowing to secure input space regions in which the model exhibits higher trustworthiness. 
Unfortunately, this ambition is impaired by the absence of uncertainty ground truth to perform this evaluation. 
This paper examines two types of uncertainty ground truth. 
We establish the validity of those in terms of their ability to yield input space regions where a classifier can be safely used at a level that can be chosen by the end-user. 
In addition, we investigate two metrics that can be computed solely using a classical test set while being tied to each identified ground truth scoring function. 
We prove that the class of functions that almost surely rank inputs in the same way each ground truth are the maximizers of one of the metrics. 
Finally, we also prove that the safer input space regions spanned by those maximizers exhibit a form of optimality in the sense that no other scoring with identical properties can encompass a larger volume of correct predictions. 
Our claims are also supported by empirical evidence in various scenarios. 
We hope that our contributions will unlock a wider use of UQ methods for safer machine learning practices. 

A potential misuse of those metrics would arise from not taking into account finite sample randomness in the reported metric values. 
Note that since the investigated metrics rely on well studied performance measures (ROC-AUC and C-index), statistical tests designed to assess the significance of performance gaps can be readily used for their uncertainty analogues.

\subsubsection*{Acknowledgements}
The authors would like to thank Geneviève Robin, Jean-Philippe Vert and Alex Nowak for the valuable discussions and insightful feedback that helped shaping the ideas presented in this work. We are also thankful to Thomas Chaigneau
for its valuable help on the OASST2's experiments.

\bibliographystyle{plainnat}
\bibliography{article.bib}

\section*{Checklist}



 \begin{enumerate}

 \item For all models and algorithms presented, check if you include:
 \begin{enumerate}
   \item A clear description of the mathematical setting, assumptions, algorithm, and/or model. [Yes Applicable]
   \item An analysis of the properties and complexity (time, space, sample size) of any algorithm. [Not Applicable]
   \item (Optional) Anonymized source code, with specification of all dependencies, including external libraries. [Yes]
 \end{enumerate}

 \item For any theoretical claim, check if you include:
 \begin{enumerate}
   \item Statements of the full set of assumptions of all theoretical results. [Yes]
   \item Complete proofs of all theoretical results. [Yes]
   \item Clear explanations of any assumptions. [Yes]     
 \end{enumerate}

 \item For all figures and tables that present empirical results, check if you include:
 \begin{enumerate}
   \item The code, data, and instructions needed to reproduce the main experimental results (either in the supplemental material or as a URL). [Yes]
   \item All the training details (e.g., data splits, hyperparameters, how they were chosen). [Yes]
         \item A clear definition of the specific measure or statistics and error bars (e.g., with respect to the random seed after running experiments multiple times). [Yes]
         \item A description of the computing infrastructure used. (e.g., type of GPUs, internal cluster, or cloud provider). [Yes]
 \end{enumerate}

 \item If you are using existing assets (e.g., code, data, models) or curating/releasing new assets, check if you include:
 \begin{enumerate}
   \item Citations of the creator If your work uses existing assets. [Not Applicable]
   \item The license information of the assets, if applicable. [Not Applicable]
   \item New assets either in the supplemental material or as a URL, if applicable. [Not Applicable]
   \item Information about consent from data providers/curators. [Not Applicable]
   \item Discussion of sensible content if applicable, e.g., personally identifiable information or offensive content. [Not Applicable]
 \end{enumerate}

 \item If you used crowdsourcing or conducted research with human subjects, check if you include:
 \begin{enumerate}
   \item The full text of instructions given to participants and screenshots. [Not Applicable]
   \item Descriptions of potential participant risks, with links to Institutional Review Board (IRB) approvals if applicable. [Not Applicable]
   \item The estimated hourly wage paid to participants and the total amount spent on participant compensation. [Not Applicable]
 \end{enumerate}

 \end{enumerate}


\newpage
\onecolumn

\aistatstitle{Legitimate ground-truth-free metrics for deep uncertainty classification scoring: \\
Supplementary Materials}

\section{UQ METHOD OVERVIEW}
\label{ap:uq_overview}

One of the goals of the present paper is to shed light on well behaved metrics to compare UQ methods. 
As such, this appendix reviews some of the most popular methods in this field. 
Since $\hat{\boldsymbol\pi}$ is a softmax vector of probabilities, the neural network can be regarded as an uncertainty representation of its own. 
UQ scorings computed from softmax probabilities constitute a baseline studied in \citep{hendrycks2016baseline}. 
Those scoring functions are presented in \cref{sub:scoring_functions}.

The softmax baseline offers very limited additional insights compared to the network output. 
In binary classification, $\hat{\pi}_1$ is the predicted probability for class 1 given $\mathbf{x}$ and is a sufficient statistic of the corresponding Bernoulli distribution. 
Scoring functions depending on $\hat{\pi}_1$ thus capture the same kind of information with respect to the uncertainty carried by that Bernoulli distribution. 
To gain additional insights, authors rapidly moved to a more flexible uncertainty representation involving a hierarchical model sometimes also referred to as second order probabilities. 
In classification, the set of first order distributions are multinomial/categorical distributions and live in the standard $\ell$-dimensional simplex where $\ell$ is the cardinal of $\mathcal{Y}$. 
A second order distribution should allocate probabilities to each element of the simplex which is precisely what the Dirichlet distribution does.

\citet{sensoy2018evidential} introduced a model called evidential network (EvN) whose output is a vector $\mathbf{e} \left( \mathbf{x} \right) \in {\mathbb{R}^{+}}^{\ell} $ such that $ \hat{\Pi}|\mathbf{x}\sim \textrm{Dir}\left( \mathbf{e} + \mathbf{1} \right)$, with a specific architecture and objective loss in the form of an integrated risk with respect to $\hat{\boldsymbol\pi}$. 
Dirichlet/Mutinomial hierarchical models is a typical case where Bayesian statistics offer the possibility to compute the posterior distribution $p_{\hat{\Pi}|\mathbf{x}, \mathcal{D}}$ where $\mathcal{D}$ are observed training data. 
Usually, since $\hat{\boldsymbol\pi} \left( \mathbf{x} \right)$ is the first level multinomial distribution, the randomness in $\hat{\Pi}|(\mathbf{x},\mathcal{D})$ can be viewed as arising from the randomness of $\Theta|\mathcal{D}$, where $\Theta$ denotes the model parameters. 
If training data would allow to find out that $p_{\Theta|\mathcal{D}} = \delta_{\boldsymbol\theta}$ then the (epistemic) uncertainty in $\hat{\Pi}$ would be wiped out. 
Inferring the posterior parameter distribution is precisely what Bayesian neural networks (BNNs) \citep{blundell2015weight,graves2011practical} propose to do. 

Another very popular UQ technique is called deep ensembles (DEs) and was introduced by~\citep{lakshminarayanan2017simple}. 
In this case, the idea is to use multiple instances $\left( \hat{\boldsymbol\pi}^{(u)} \right)_{u=1}^{n_{\textrm{ens}}} $ of a single network architecture. 
The finite set of learnt functions are different due to random sampling of the initial network parameters, shuffling during stochastic gradient descent and adversarial perturbations. 
DEs share some similarity with BNNs as, in practice, samples from the model posterior are used which de facto constitute an ensemble of models.

Another way to introduce a distribution over the network parameters is to view each of them as a Bernoulli variable allowing each parameter to switch between its learnt value and zero, i.e. through dropout~\citep{srivastava2014dropout}. 
By sampling from these Bernoulli variables, we obtain MC Dropout~\citep{gal2016dropout} and a set of predictive probabilities is also issued. 

For end-users interested in the OOD detection use case, it is very instrumental to leverage input distances as shown by \citet{liu2020simple}. The authors use a Gaussian process on logits whose kernel brings distance awareness and induces second order probabilities through the softmax function. 

Instead of sampling by using multiple models, i.e. through the distribution of $\Theta$, another possibility is to do this through the input distribution by leveraging augmentation mechanisms. 
\citet{wang2019aleatoric} propose to do so at test time, hence the name of this method called test time augmentation (TTA). 
The data augmentation being a random process, it induces a distribution on $\hat{\Pi} | \mathbf{x}$.

Finally, it should also be mentioned that another branch of the literature is dedicated to UQs learnt from a held-out set of data iid under $p_{X,Y}$. 
One can typically learn a ``meta-classifier'' to map inputs to their likeliness of being incorrectly classified by $\hat{\boldsymbol\pi}$, see~\citep{qiu2022detecting} and references therein for examples of such approaches. 
Here again, the meta-classifier can be chosen to output first or second order types of representation.

\section{SCORING FUNCTIONS} 
\label{sub:scoring_functions}

As evoked in \cref{ap:uq_overview}, in binary classification, when quantifying uncertainty using first order probabilities, the scoring functions one can craft all carry roughly the same information. 
Let $\tilde{Y}|\mathbf{x}$ denotes a random variable with distribution $\textrm{Ber} \left( \hat{\pi}_1 \right) $. 
Possible choices for a scoring $s \left( \mathbf{x} \right) $ are typically $\textrm{Var} \left[ \tilde{Y}| X=\mathbf{x} \right] $, the entropy $H\left(\tilde{Y}| X=\mathbf{x} \right) $ or the gap $1 - \hat{\pi}_{\hat{y}}$.

For uncertainty representation leveraging second order probabilities, there is increased flexibility in the design of scoring functions.
Concerning BNNs, DEs and MCD, one has access to a set of predicted softmax vectors $ \left( \hat{\boldsymbol\pi}^{(u)} \right)_{u=1}^{n_{\textrm{ens}}} $. 
Usual scoring functions are defined as follows:
\begin{itemize}
    \item Variation Ratio $ \textrm{VR} = 1 - \hat{\pi}_{\hat{y}}$,
    \item total entropy $ \textrm{TE} = H \left( \tilde{Y}^{(\textrm{ens})}|X=\mathbf{x} \right) $ where $\tilde{Y}^{(\textrm{ens})}|X=\mathbf{x}$ denotes a random variable whose distribution is $\hat{\boldsymbol\pi}^{(\textrm{ens})} = \frac{1}{n_{\textrm{ens}}} \sum_{u=1}^{n_{\textrm{ens}}}  \hat{\boldsymbol\pi}^{(u)}$,
    \item aleatoric entropy $\textrm{AE} = \frac{1}{n_{\textrm{ens}}} \sum_{u=1}^{n_{\textrm{ens}}} H \left( \tilde{Y}^{(u)}|X=\mathbf{x} \right) $ where $\tilde{Y}^{(u)}|X=\mathbf{x}$ denotes a random variable whose distribution is $\hat{\boldsymbol\pi}^{(u)}$,
    \item mutual information $\textrm{MI}$ which by definition is equal to $\textrm{TE}-\textrm{AE}$,
    \item and free energy $-T \cdot \log \sum_{y\in \mathcal{Y}} e^{c_y/T}$ where $\mathbf{c}$ is the vector of logits (pre-softmax output) and $T$ is a temperature parameter \citep{liu2020energy}.
\end{itemize}
As can be understood from the above nomenclature, ensemble based UQ allows finer grained scoring which can disambiguate aleatoric from epistemic uncertainty. 
Concerning TTA, one uncertainty score proposed by the authors is obtained by computing the entropy of the empirical multinomial distribution obtained from the set of predicted labels to which augmented inputs were mapped. 
According to the authors, this score should align with aleatoric uncertainty as the process simulates outcomes in the vicinity of an input. 
For epistemic uncertainty, the authors use MCD without TTA. 


\section{METRICS ASSESSING RANKING PERFORMANCE VERSUS METRICS ASSESSING CALIBRATION PERFORMANCE} 
\label{sec:metrics_assessing_ranking_performance_versus_calibration_performance}
\begin{figure*}[!h]
  \centering
    \includegraphics[width=.4\textwidth]{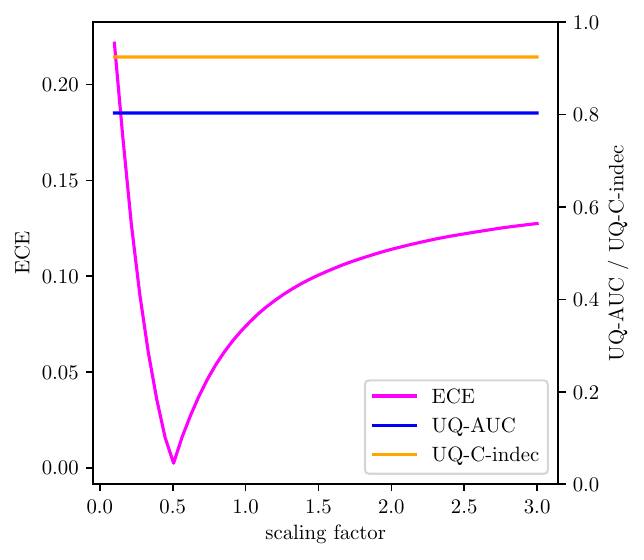}
  \caption{ECE, UQ-AUC and UQ-C-index as functions of the scaling factor of Temperature Scaling.\label{fig:ece}}
\end{figure*}

The scope of this paper is input ranking based on classification error risk. 
Equipped with a well-behaved ranking, one can secure regions of the input space where the model, whose uncertainty is being quantified, can be used safely. 
As an illustration of the discrepancy between this scope and calibration, we designed an illustrative example leveraging Temperature Scaling \citep{guo2017calibration}. 
If the Temperature Scaling transform is applied to the softmax probabilities issued by $\hat{\pi}$ with different scaling factors, then different ECE values are observed. 
If inputs are ranked using the softmax scoring function, then applying Temperature Scaling will not modify the ranking. 
In this case, metrics such as UQ-AUC or UQ-C-index are invariant to this transform as shown in the following figure obtained using the synthetic dataset as can be seen on \cref{fig:ece}.

This simple example shows that very high UQ-AUC or UQ-C-index values can be achieved by the softmax uncertainty scoring (thereby allowing to correctly identify risk-prone inputs) while the underlying softmax probabilities are not necessarily well calibrated. 

To further emphasize that calibration do not capture the ground truth signal sought in this paper, one can check the correlation of ECE and $\kappa (s, \phi)$ or $\kappa (s, \varphi)$. 
On the synthetic dataset, the achieved correlations are respectively equal to 0.09\% and -10.57\%.

Besides, in \cref{sub:optimality_of_functions_}, we explain that \eqref{eq:nested_sets_phi} holds for infinite families of scorings while calibration holds for a unique score. 
This statement refers to \citep[eq. (1)]{guo2017calibration} defining perfect calibration of a confidence estimate taking values in the full range of the $[0;1]$ interval. 
Two estimates achieving perfect calibration in this sense need to be equal almost everywhere. 
This is in sharp contrast to the elements of $\mathcal{E}_{\phi}$ which do not have to be equal almost everywhere while all maximizing UQ-AUC. 
This is why we argue that ECE and its variants solve a different evaluation problem compared to the metrics investigated in this paper.


\section{PROOF OF \cref{thm:proof_nested_sets_phi}}\label{ap:proof_nested_sets_phi}

\cref{thm:proof_nested_sets_phi} shows that sub-level sets $\mathcal{L}_\beta \left( s \right) := \left\{ \mathbf{x} \in \mathcal{X} \middle| s \left( \mathbf{x} \right) \leq \beta \right\}$ of any scoring in the equivalence class of ground truth $\phi$ yield a family of nested sets in which the accuracy of the predictor $\hat{y}$ decreases with $\beta$. 
To prove this result, we will need to leverage an intermediate result which is first proved in the following paragraphs. 

\subsection{Intermediate result}

\begin{theorem}\label{thm:interm_result}
    For any pair of scoring function $s_1$ and $s_2 \in \mathcal{E}_{s}$ and any $\beta \in \mathbb{R}$, $\exists\gamma\in \mathbb{R}$ such that 
    \begin{equation}
        P \left( X \in \mathcal{L}_\beta \left( s_1 \right) \setminus \mathcal{L}_\gamma \left( s_2 \right) \cup  \mathcal{L}_\gamma \left( s_2 \right) \setminus \mathcal{L}_\beta \left( s_1 \right)   \right) = 0.\label{eq:level_set_equi}
    \end{equation}
\end{theorem}
By definition of $\mathcal{E}_{s}$, we have
\begin{align}
     \Leftrightarrow & \begin{cases} P \left( \left\{  s \left( X \right) \geq s \left( X' \right)   \right\} \cap \left\{  s_1 \left( X \right) < s_1 \left( X' \right)  \right\}   \right) &=0 \\ P \left(  \left\{  s \left( X \right) < s \left( X' \right)   \right\} \cap \left\{  s_1 \left( X \right) \geq s_1 \left( X' \right)  \right\}   \right)&=0  \end{cases}.\label{eq:thm_phi}
\end{align}

If $s$ is constant $p_X$-almost everywhere, then $P \left( X \in \mathcal{L}_\beta \left( s \right) \right) = 0 $ or 1. 
This implies that $P \left(  s \left( X \right) < s \left( X' \right)  \right) = 0$.
Leveraging the top equation in \eqref{eq:thm_phi}, one can write
\begin{align}
    & P \left( \left\{  s \left( X \right) \geq s \left( X' \right)   \right\} \cap \left\{  s_1 \left( X \right) < s_1 \left( X' \right)  \right\}   \right) =0,\\
    \Leftrightarrow & P \left( \left\{  s \left( X \right) \geq s \left( X' \right)   \right\} \cap \left\{  s_1 \left( X \right) < s_1 \left( X' \right)  \right\}   \right) + P \left(  s \left( X \right) < s \left( X' \right)  \right)  =0,\\
    \Leftrightarrow & P \left( \left\{  s \left( X \right) \geq s \left( X' \right)   \right\} \cap \left\{  s_1 \left( X \right) < s_1 \left( X' \right)  \right\}   \right) + P \left( \left\{ s \left( X \right) < s \left( X' \right)  \right\} \cap \left\{  s_1 \left( X \right) < s_1 \left( X' \right)  \right\}  \right)  =0,\\
    \Leftrightarrow & P \left(  s_1 \left( X \right) < s_1 \left( X' \right) \right) =0.
\end{align}

This means that $s_1$ is also constant $p_X$-almost everywhere. 
If $P \left( X \in \mathcal{L}_\beta \left( s \right) \right) = 0 $, it suffices to choose $\gamma$ sufficiently small so that $P \left( X \in \mathcal{L}_\gamma \left( s_1 \right) \right) = 0 $ as well and \eqref{eq:level_set_equi} holds for the scorings $s$ and $s_1$. 
Likewise, if $P \left( X \in \mathcal{L}_\beta \left( s \right) \right) = 1 $, it suffices to choose $\gamma$ sufficiently high so that $P \left( X \in \mathcal{L}_\gamma \left( s_1 \right) \right) = 1 $ as well and \eqref{eq:level_set_equi} holds for the scorings $s$ and $s_1$.

We now proceed with a scoring $s$ that is not constant $p_X$-almost everywhere and thus $P \left(  s \left( X \right) < s \left( X' \right)  \right)>0$. Based on the top equation in \eqref{eq:thm_phi}, this implies that 
\begin{align}
   & P \left(  s_1 \left( X \right) \geq s_1 \left( X' \right) |  s \left( X \right) < s \left( X' \right)  \right) = 0,\\
   \Leftrightarrow & P \left(  s_1 \left( X \right) < s_1 \left( X' \right) |  s \left( X \right) < s \left( X' \right)  \right) = 1.
\end{align}

Now suppose $X \in \mathcal{L}_\beta \left( s \right)$ and $X' \notin \mathcal{L}_\beta \left( s \right)$. 
Since this implies that $s \left( X \right) < s \left( X' \right) $, we deduce
\begin{align}
    P \left(  s_1 \left( X \right) < s_1 \left( X' \right) |  X \in \mathcal{L}_\beta \left( s \right), X' \notin \mathcal{L}_\beta \left( s \right)  \right) = 1.\label{eq:level_sep}
\end{align}
The above equation means that $\mathcal{L}_\beta \left( s \right)$ and its complementary set define a partition of the input space such that $s_1$ almost surely takes values smaller in $\mathcal{L}_\beta \left( s \right)$ than in its complementary set. 
Define $\gamma$ as the following essential supremum:
\begin{align}
    \gamma := \inf \left\{ M \in \mathbb{R} | P \left( X \in \mathcal{L}_\beta \left( s \right) | X \notin \mathcal{L}_M\left( s_1  \right)   \right) =0 \right\}
\end{align}

By construction, we have $ P \left( \mathcal{L}_{\beta}\left( s \right) \setminus \mathcal{L}_\gamma\left( s_1  \right) \right)=0 $.
Moreover, thanks to \eqref{eq:level_sep}, we know that $ P \left( \mathcal{L}_\gamma\left( s_1  \right) \setminus \mathcal{L}_{\beta}\left( s \right) \right)=0 $
otherwise there would be a subset $U \subset \mathcal{L}_\gamma\left( s_1  \right)$ in the complementary set of $\mathcal{L}_{\beta}\left( s \right) $ such that $P \left( X \in U \right) >0 $ and one could find a smaller scalar than $\gamma$ allowing to exclude $U$ while keeping $P \left( X \in \mathcal{L}_\beta \left( s \right) | X \notin \mathcal{L}_{\gamma}\left( s_1  \right) \setminus U  \right) =0$. 
Relation \eqref{eq:level_set_equi} thus holds for scoring $s$ and $s_1$.

Since Relation \eqref{eq:level_set_equi}  is symmetric and also holds for scoring $s$ and $s_2$, by transitivity, it holds as well for $s_1$ and $s_2$.

\subsection{Main proof}
In this proof, we are interested in the following quantities
\begin{align}
    E \left( \beta; s \right) := P \left( Y \neq \hat{Y} | X \in \mathcal{L}_{\beta} \left( s \right) \right).
\end{align}

Remembering that  $\phi\left( \mathbf{x} \right) = P \left(   Y \neq \hat{Y} | X= \mathbf{x}\right) $, one can write
\begin{align}
    E \left( \beta; \phi \right) &= \frac{\int_{\mathcal{L}_{\beta} \left( \phi \right)  } P \left(   Y \neq \hat{Y} | X=\mathbf{x}\right) p_X \left( \mathbf{x} \right) d\mathbf{x}}{P \left( X \in \mathcal{L}_{\beta} \left( \phi \right)  \right) },\\
    & \leq \frac{\int_{\mathcal{L}_{\beta} \left( \phi \right)} \beta p_X \left( \mathbf{x} \right) d\mathbf{x}}{P \left( X \in \mathcal{L}_{\beta} \left( \phi \right)  \right) },\\
    & \leq \beta.\label{eq:bounded_level_set_acc}
\end{align}

Besides, if $\phi$ is almost surely constant, then relation \eqref{eq:nested_sets_phi} is trivially met\footnote{In this case, the sub-level sets $\mathcal{L}_{\beta} \left( \phi \right) $ have a measure either equal to 0 or 1. Only those with a strictly positive measure are involved in \eqref{eq:nested_sets_phi} and then $E \left( \beta; \phi \right) = P \left( Y \neq \hat{Y}  \right) = 1 - \textrm{ACC} $.}. 
Let us thus now suppose that this co-domain contains at least two distinct elements $\beta$ and $\beta+h$ such that $\beta < 1$ and where $h$ is a positive scalar, smaller than $1 - \beta$. 
One can write
\begin{align}
    E \left( \beta + h; \phi \right) &= \frac{P \left( Y \neq \hat{Y} , X \in \mathcal{L}_{\beta+h} \left( \phi \right)  \right) }{\underbrace{P \left( X \in \mathcal{L}_{\beta} \left( \phi \right) \right)}_{b} + \underbrace{P \left( X \in \mathcal{L}_{\beta+h} \left( \phi \right) \setminus \mathcal{L}_{\beta} \left( \phi \right)  \right) }_{a}},
\end{align}
where $a$ and $b$ are positive scalars.

The numerator of $E \left( \beta + h; \phi \right) $ can be written as
\begin{align}
    P \left( Y \neq \hat{Y} , X \in \mathcal{L}_{\beta+h} \left( \phi \right)  \right) &= \int_{\mathcal{L}_{\beta+h} \left( \phi \right)} P \left(   Y \neq \hat{Y} | X=\mathbf{x}\right) p_X \left( \mathbf{x} \right) d\mathbf{x},\\
    &= b E \left( \beta \right) + \int_{\mathcal{L}_{\beta+h} \left( \phi \right) \setminus  \mathcal{L}_{\beta} \left( \phi \right)  } P \left(   Y \neq \hat{Y} | X=\mathbf{x}\right) p_X \left( \mathbf{x} \right) d\mathbf{x}.
\end{align}
We also know that for any $\mathbf{x} \in \mathcal{L}_{\beta+h} \left( \phi \right) \setminus \mathcal{L}_{\beta} \left( \phi \right) $, $P \left(   Y \neq \hat{Y} | \mathbf{x}\right) > \beta$. We thus can write

\begin{align}
    E \left( \beta + h; \phi \right) &>\frac{b E \left( \beta ; \phi\right) + a \beta}{b + a}
\end{align}

We can now leverage relation \eqref{eq:bounded_level_set_acc} stating that $E \left( \beta; \phi \right)  \leq \beta$ to obtain

\begin{align}
    E \left( \beta + h ; \phi\right) &> \frac{b E \left( \beta ; \phi\right) + a E \left( \beta; \phi \right)}{b + a},\\
    &> E \left( \beta; \phi \right).
\end{align}
We hereby conclude that $E \left( . ; \phi \right) $ is a strictly increasing function of $\beta$. 
An illustration compliant with claim is shown in \cref{fig:risk_control}. 
By observing that $E \left( \beta=0^+; \phi \right)=0 $ and $E \left( \beta=1; \phi \right)=P \left( Y\neq\hat{Y} \right) = 1 - \text{ACC}$, we know that the set $\mathscr{C}_{\text{mce}} \left( \mathcal{E}_\phi \right) $ of achievable values of $E \left(. ; \phi \right) $ is a subset of $ \left] 0; 1-\textrm{ACC} \right]$. 
Let $\psi_{\phi} = E \left( . ; \phi \right)^{-1} $, then obviously $E \left(\psi_{\phi} \left( \gamma \right)  ; \phi \right) = P \left( Y \neq \hat{Y} | X \in \mathcal{L}_{\psi_{\phi} \left( \gamma \right)  } \left( s \right) \right) = \gamma$, $\forall\gamma \in \mathscr{C}_{\text{mce}} \left( \mathcal{E}_\phi \right) $ and \eqref{eq:nested_sets_phi} holds for ground truth scoring $\phi$. 

Moreover, thanks to \cref{thm:interm_result}, we know that $\forall s \in \mathcal{E}_\phi$ there is bijection $\psi_s$ such that $E \left( \psi_s \left( \gamma \right)  , s \right) =  E \left( \gamma, \psi_{\phi}^{-1} \circ \phi \right) = \gamma$. 
As $\psi_s$ is clearly a strictly increasing function of $\gamma$, then \eqref{eq:nested_sets_phi} holds for $s$ as well which concludes the proof.

\section{PROOF OF \cref{thm:proof_nested_sets_varphi}}
\label{ap:proof_nested_sets_varphi}

\cref{thm:proof_nested_sets_varphi} shows that sub-level sets $\mathcal{L}_\beta \left( s \right) $ of any scoring in the equivalence class of ground truth $\varphi$ yield a family of nested sets in which the probability of predicting the same class label as the Bayes classifier decreases with $\beta$. As such, we are interested in this proof in the following quantities
\begin{align}
    E \left( \gamma; s \right) := P \left( \hat{Y}_{\textrm{bay}} \neq \hat{Y} | X \in \mathcal{L}_{\gamma} \left( s \right) \right).
\end{align}

Remembering that $\varphi(\mathbf{x}) = \frac{1}{2} \Vert f_\theta(\mathbf{x}) - \boldsymbol\delta_{\hat{y}_{\textrm{bay}}}\Vert_1 = 1 - \hat{\pi}_{\hat{y}_{\text{bay}}}$, one can write
\begin{align}
    P \left( \hat{Y}_{\textrm{bay}}\neq \hat{Y}, X \in \mathcal{L}_{\beta} \left( \varphi \right)  \right) &= \int_{\mathcal{L}_{\beta} \left( \varphi \right) }  P \left( \hat{Y}_{\textrm{bay}}\neq \hat{Y}| X=\mathbf{x} \right)p_X \left( \mathbf{x} \right) d\mathbf{x},\\
    &= \int_{\mathcal{L}_{\beta} \left( \varphi \right) } \mathds{1}_{ \varphi \left( \mathbf{x} \right) \geq \sfrac{1}{2} }\: p_X \left( \mathbf{x} \right) d\mathbf{x},\\
    &= \begin{cases}
    0 & \textrm{if } \beta \leq \sfrac{1}{2},\\
    \int_{\mathcal{L}_{\beta} \left( \varphi \right) \setminus \mathcal{L}_{\sfrac{1}{2}} \left( \varphi \right)  }p_X \left( \mathbf{x} \right) d\mathbf{x} & \textrm{otherwise}
    \end{cases},\\
    &= \max\left(0,  P \left( X \in \mathcal{L}_{\beta} \left( \varphi \right)  \right)- P \left( X \in \mathcal{L}_{\sfrac{1}{2}} \left( \varphi \right)  \right)\right) .
\end{align}

We deduce that $E \left( \beta; \varphi \right) = P \left( \hat{Y}_{\textrm{bay}}\neq \hat{Y} | X \in \mathcal{L}_{\beta} \left( \varphi \right) \right) = \max \left( 0, 1- \frac{P \left( X \in \mathcal{L}_{\sfrac{1}{2}} \left( \varphi \right) \right)}{P \left( X \in \mathcal{L}_{\beta} \left( \varphi \right)  \right)} \right)  $ which implies that $E \left( .; \varphi \right) $ is null on $\left] 0;\sfrac{1}{2} \right] $ and increasing on $\left] \sfrac{1}{2};1 \right] $. 
An illustration compliant with claim is shown in \cref{fig:risk_control}. 
In particular, $E \left( 1; \varphi  \right) = 1-P \left( X \in \mathcal{L}_{\sfrac{1}{2}} \left( \varphi \right)  \right) = P \left( \hat{Y}_{\textrm{bay}}\neq \hat{Y} \right)$. 
Consequently,  we know that the set $\mathscr{C}_{\text{mbc}} \left( \mathcal{E}_\varphi \right) $ of achievable values of $E \left( . ; \varphi \right) $ is a subset of $ \left] 0; P \left( \hat{Y}_{\textrm{bay}}\neq \hat{Y} \right) \right]$.
Let $\psi_{\varphi} = E \left( . ; \varphi \right)^{-1} $, then obviously $E \left(\psi_{\varphi} \left( \gamma \right)  ; \varphi \right) = P \left( \hat{Y}_{\textrm{bay}} \neq \hat{Y} | X \in \mathcal{L}_{\psi_{\varphi} \left( \gamma \right)  } \left( s \right) \right) = \gamma$, $\forall\gamma \in \mathscr{C}_{\text{mbc}} \left( \mathcal{E}_\varphi \right) $ and \eqref{eq:nested_sets_varphi} holds for ground truth scoring $\varphi$. 

Moreover, thanks to \cref{thm:interm_result}, we know that $\forall s \in \mathcal{E}_\varphi$ there is bijection $\psi_s$ such that $E \left( \psi_s \left( \gamma \right)  , s \right) =  E \left( \gamma, \psi_{\varphi}^{-1} \circ \varphi \right) = \gamma$. 
As $\psi_s$ is clearly a strictly increasing function of $\gamma$, then \eqref{eq:nested_sets_varphi} holds for $s$ as well which concludes the proof.

\section{PROOF OF \cref{thm:uq_auc}}\label{ap:proof_thm_uq_auc}
This appendix contains a proof that $s$ is a maximizer of UQ-AUC iff $s\in \mathcal{E}_\phi$. 
The fact that $\phi$ and any increasing measurable function of $\phi$ is a maximizer of UQ-AUC is a direct consequence of~\citep[Proposition 1]{Clemencon2008ranking} and this part of the proof is thus omitted. 
It remains to prove that other elements in $\mathcal{E}_\phi$ are also maximizers and that all maximizers belong to $\mathcal{E}_\phi$.

We start by showing that any $s\in \mathcal{E}_\phi$ is a maximizer of UQ-AUC. 
To comply with notations in \citep{Clemencon2008ranking}, let us introduce the following binary relation
\begin{align}
    r_s\left( \mathbf{x},\mathbf{x}' \right) = \begin{cases} 1 & \text{if } s \left( \mathbf{x} \right) \geq s \left( \mathbf{x}' \right)  \\ -1 & \text{otherwise}  \end{cases}.
\end{align}
Let us also define the following random variables $V   = \mathds{1}_{ Y \neq \hat{Y} | X }$, $ V'  = \mathds{1}_{ Y' \neq \hat{Y}'|X' }$ and $Z = V - V'$. 
\citep{Clemencon2008ranking} proved that $s$ is a maximizer of the UQ-AUC iff it minimizes
\begin{align}
    P \left( Z \cdot r_s\left( X,X' \right) < 0  \right) &= P \left(\left\{ Z \cdot r_s\left( X,X' \right) < 0 \right\} \cap \left\{ Z \cdot r_\phi\left( X,X' \right) < 0 \right\}  \right)
\end{align}
By definition of $\mathcal{E}_{\phi}$, we also have $P \left( r_\phi\left( X,X' \right)  \neq r_s\left( X,X' \right) \right)=0 $ and events $\left\{ Z \cdot r_s\left( X,X' \right) < 0 \right\}$ and $\left\{ Z \cdot r_\phi\left( X,X' \right) < 0 \right\} $ have the same probability therefore $s$ is also a maximizer of UQ-AUC. 

We now move to the proof that any maximizer of UQ-AUC is in $\mathcal{E}_\phi$.
Let $s$ denote a maximizer of UQ-AUC such that $\exists \mathcal{A}\subset \mathcal{X}^2$ with $\lambda = P \left( \left( X, X' \right)\in  \mathcal{A}\ \right) >0 $ and $P \left( r_{\phi} \left( X, X' \right)\cdot  r_{s} \left( X, X' \right)  <0 |  \left( X, X' \right)\in  \mathcal{A} \right) = 1$. 

Let us then define the following binary relation
\begin{align}
    \tilde{r}\left( \mathbf{x},\mathbf{x}' \right) = \begin{cases} \max \left\{ r_s\left( \mathbf{x},\mathbf{x}' \right) ; r_\phi\left( \mathbf{x},\mathbf{x}' \right) \right\} & \text{if } Z>0 \text{ and } \left( \mathbf{x},\mathbf{x}' \right)\in \mathcal{A} \\ \min \left\{ r_s\left( \mathbf{x},\mathbf{x}' \right) ; r_\phi\left( \mathbf{x},\mathbf{x}' \right) \right\} & \text{if } Z<0 \text{ and } \left( \mathbf{x},\mathbf{x}' \right)\in \mathcal{A}  \\ r_\phi\left( \mathbf{x},\mathbf{x}' \right) & \text{otherwise}  \end{cases}.
\end{align}

One then has 
\begin{align}
    P \left( Z \cdot \tilde{r}\left( X,X' \right) < 0 \right) &=  P \left( Z \cdot \tilde{r}\left( X,X' \right) < 0 , \left( \mathbf{x},\mathbf{x}' \right)\in \mathcal{A} \right) +  P \left( Z \cdot r_\phi\left( X,X' \right) < 0 , \left( \mathbf{x},\mathbf{x}' \right)\notin \mathcal{A} \right).
\end{align}
Because inside $\mathcal{A}$ we know that $r_s\left( X,X' \right) = -r_{\phi}\left( X,X' \right)$, by construction of $\tilde{r}$, we have $P \left( Z \cdot \tilde{r}\left( X,X' \right) < 0 , \left( \mathbf{x},\mathbf{x}' \right)\in \mathcal{A} \right)=0$. 
This implies that $P \left( Z \cdot \tilde{r}\left( X,X' \right) < 0 \right) =  P \left( Z \cdot r_\phi\left( X,X' \right) < 0 , \left( \mathbf{x},\mathbf{x}' \right)\notin \mathcal{A} \right) <  P \left( Z \cdot r_\phi\left( X,X' \right) < 0 \right)$. 
This is in contraction with $\phi$ being a maximizer of UQ-AUC which concludes the proof.

\section{PROOF OF \cref{thm:phi_optim}}\label{ap:proof_phi_optim}
In this appendix, one aims at showing that a scoring $s$ for which \eqref{eq:nested_sets_phi} holds is not uniformly better than $s^* \in \mathcal{E}_s$. 
In this proof, we will use the following random variables $V \left( \mathbf{x} \right)  = \mathds{1}_{ Y \neq \hat{Y} | \mathbf{x} }$ and $ V'\left( \mathbf{x}' \right)  = \mathds{1}_{ Y' \neq \hat{Y}'|\mathbf{x}'}$. 
For the sake concision, we also define $V=V \left( X \right) $, $V'=V' \left( X' \right) $, $S = s \left( X \right) $ and $S' = s \left( X' \right) $.

We will also be need to introduce the scoring $\tilde{s} = \psi_s^{-1}\circ s$ where $\psi_s$ is the strictly increasing function allowing to re-index the sub-level set of $s$ to achieve \eqref{eq:nested_sets_phi}. 
For any $\gamma \in \mathscr{C}_{\text{mbc}} \left( \mathcal{E}_s \right)$, we thus have
\begin{align}
    \mathbf{x} \in \mathcal{L}_{\gamma} \left( \tilde{s} \right) &\Leftrightarrow \tilde{s} \left( \mathbf{x} \right) \leq \gamma \\
    & \Leftrightarrow  \psi_s \left(   \tilde{s} \left( \mathbf{x} \right) \right) \leq \psi_s \left( \gamma \right) \\
    & \Leftrightarrow s \left( \mathbf{x}  \right)   \leq \psi_s \left( \gamma \right) \\
    & \Leftrightarrow \mathbf{x} \in \mathcal{L}_{\psi_s \left( \gamma \right) } \left( s \right).
 \end{align}
Since $\psi_s^{-1}$ is an increasing function, $\tilde{s} \in \mathcal{E}_s$ and \eqref{eq:nested_sets_phi} also holds for this scoring. 
This means that the bijection to re-index sub-level sets of $\tilde{s} $ as part of property \eqref{eq:nested_sets_phi} is the identity function. 
Likewise, we define $\tilde{s}^*$ from $s^*$ in the same way. 

Suppose $s^* \prec_{\text{mce}} s \Leftrightarrow \tilde{s}^* \prec_{\text{mce}} \tilde{s} $. 
The UQ-AUC achieved by some scoring $s$ then writes
\begin{align}
    \textrm{UQ-AUC}\left( s \right) &= P \left( S \leq  S' | V  < V'    \right),\\
    &= P \left( \tilde{S} \leq  \tilde{S}' | V  < V'    \right),\\
    &= \int P \left( \tilde{S} \leq  \tilde{S}', \tilde{S}'=\gamma | V  < V'\right) d\gamma,\\
    &= \int P \left( \tilde{S} \leq  \gamma | V  < V'\right) d\gamma,\\
    &= \int P \left( X \in  \mathcal{L}_\gamma \left( \tilde{s} \right)  | V  < V' \right) d\gamma,\\
    &= \int P \left( X \in  \mathcal{L}_\gamma \left( \tilde{s} \right)  | V=0  , V'=1 \right) d\gamma,\\
    &= \int P \left( X \in  \mathcal{L}_\gamma \left( \tilde{s} \right)  | V=0 \right) d\gamma,\\
    &= \int P \left( X \in  \mathcal{L}_\gamma \left( \tilde{s} \right)  | Y = \hat{Y} \right) d\gamma,\\
    &> \int P \left( X \in  \mathcal{L}_{ \gamma  } \left( \tilde{s}^* \right) | Y = \hat{Y} \right)  d\gamma.
    \label{eq:uq-auc-interm}
\end{align}

We now can conclude that if $s$ was dominating this $s^*$ in the sense defined in \cref{thm:phi_optim}, then it would achieve a larger UQ-AUC than the maximizers of the latter which is not possible.

\section{PROOF OF \cref{thm:uq_c_index}}
\label{ap:proof_uq_cindex_optimal_score}

This appendix addresses the question of which family of functions maximizes the uncertainty concordance index. 
We address binary classification only, i.e. $\mathcal{Y} = \left\{ 0;1 \right\}$. 
We remind that $\textrm{UQ-C-index}\left( s\right) $ refers to the asymptotical definition of this metric which reads
\begin{align}
    \textrm{UQ-C-index}\left( s \right) &= P \left( s \left( X\right) \leq  s \left( X' \right) |  \Delta\left( \hat{\boldsymbol\pi} \left( X \right) , Y\right) \leq  \Delta \left( \hat{\boldsymbol\pi} \left(X'\right), Y' \right)  \right).
\end{align}

Based on the work of \citep{Clemencon2008ranking}, we know that an optimal scoring $s^*$ is spanned by checking if $\rho_+ \left( \mathbf{x}, \mathbf{x}' \right) $ is greater (or not) than $\rho_- \left( \mathbf{x}, \mathbf{x}' \right) $ where those functions are defined as follows:
\begin{align}
    \rho_+ \left( \mathbf{x}, \mathbf{x}' \right) &= P \left( \Delta \left( \hat{\boldsymbol\pi} \left(\mathbf{x}\right), Y\right) -  \Delta \left( \hat{\boldsymbol\pi} \left(\mathbf{x}'\right), Y' \right)  >0 | X=\mathbf{x}, X'=\mathbf{x}' \right),\\
    \rho_- \left( \mathbf{x}, \mathbf{x}' \right) &= P \left( \Delta \left( \hat{\boldsymbol\pi} \left(\mathbf{x}\right), Y\right) -  \Delta \left( \hat{\boldsymbol\pi} \left(\mathbf{x}'\right), Y' \right)  <0 | X=\mathbf{x}, X'=\mathbf{x}' \right).
\end{align}

Using \eqref{eq:delta}, we thus have 
\begin{align}
    \rho_+ \left( \mathbf{x}, \mathbf{x}' \right) &= P \left( \hat{\pi}_{Y'} \left( \mathbf{x}' \right)  > \hat{\pi}_{Y}\left( \mathbf{x} \right) | X=\mathbf{x}, X'=\mathbf{x}' \right),\\
    &= \sum_y \sum_{y'}P \left( \hat{\pi}_{y'}\left( \mathbf{x}' \right) > \hat{\pi}_{y}\left( \mathbf{x} \right) | X=\mathbf{x}, X'=\mathbf{x}',Y=y,Y'=y' \right) \nonumber\\
    & \qquad\qquad\qquad\qquad\qquad\qquad\qquad\qquad\qquad\qquad p_{Y|X=\mathbf{x}} \left( y \right) p_{Y'|X'=\mathbf{x}'} \left( y' \right),\\
    &= \sum_y \sum_{y'} \mathds{1}_{ \hat{\pi}_{y'}\left( \mathbf{x}' \right)   > \hat{\pi}_{y}\left( \mathbf{x} \right)  }\: \nu_y \left(\mathbf{x} \right) \nu_{y'} \left(\mathbf{x}' \right),\label{eq:indicator_sum}
\end{align}
where $\nu_y \left(\mathbf{x} \right):= P \left( Y=y|X=\mathbf{x} \right)$ and $\nu_{y'} \left(\mathbf{x}' \right) :=  P \left( Y'=y|X'=\mathbf{x}' \right)$ are notations introduced for improved readability. 
As not all indicator functions in \eqref{eq:indicator_sum} can be equal to one simultaneously, a case by case examination leads to the following expression of $\rho_+$:
\begin{align}
    \rho_+ \left( \mathbf{x}, \mathbf{x}' \right) &= \sum_{y', y \textrm{ s.t. } \hat{\pi}_{y'}\left( \mathbf{x}' \right)   > \hat{\pi}_{y}\left( \mathbf{x} \right) }  \nu_y \left(\mathbf{x} \right) \nu_{y'} \left(\mathbf{x}' \right).\label{eq:rho_plus_cindex}
\end{align}

Since $p_{\hat{\Pi}}$ has no atom, we know that $>$ is a total order for the entries of $\hat{\boldsymbol\pi}\left( \mathbf{x} \right) $ and $\hat{\boldsymbol\pi}\left( \mathbf{x}' \right) $, i.e. there is no tie in the sorting and $\hat{\pi}_y \left( \mathbf{x} \right) \neq \hat{\pi}_{y'}\left( \mathbf{x}' \right) , \forall y, y'$. 

Let us break down the different sorting possibilities for a fixed (arbitrary) $\left( \mathbf{x}, \mathbf{x}' \right) $. 
For the sake of readability, we omit dependencies over $\mathbf{x}$ and $\mathbf{x}'$ in most of the sequel of this proof and we use $\hat{\pi}'_y$ to denote $\hat{\pi}_y \left( \mathbf{x}' \right) $ or $\hat{y}'$ to denote $\hat{y} \left( \mathbf{x}' \right) $. 
Let $\iota = \min \left( \hat{\pi}_0, \hat{\pi}_{1}, \hat{\pi}'_0,  \hat{\pi}'_{1} \right) $. 

As $\mathcal{Y} = \left\{ 0;1 \right\}$, we investigate the following cases:
\begin{itemize}
    \item $\iota = \hat{\pi}_{y_0}$, i.e. it is an entry of $\hat{\boldsymbol\pi}$ and class label with minimal probability is $y_0$. This implies that $\max \left( \hat{\pi}_0, \hat{\pi}_{1}, \hat{\pi}'_0, \hat{\pi}'_{1} \right) = \hat{\pi}_{1-y_0}$. 
    Consequently, for any $y'$ the condition under the sum in \eqref{eq:rho_plus_cindex} is always checked for $y=y_0$ and never checked for $y\neq y_0$. 
    We obtain that
    \begin{align}
        \rho_+ \left( \mathbf{x}, \mathbf{x}' \right) &= P \left( Y=y_0 | X=\mathbf{x} \right).
    \end{align}

    \item $\iota = \hat{\pi}'_{y_0}$, i.e. it is an entry of $\hat{\boldsymbol\pi}'$ and class label with minimal probability is $y_0$. This implies that $\max \left( \hat{\pi}_0, \hat{\pi}_{1}, \hat{\pi}'_0, \hat{\pi}'_{1} \right) = \hat{\pi}'_{1-y_0}$. 
    Consequently, for any $y$ the condition under the sum in \eqref{eq:rho_plus_cindex} is always checked for $y'\neq y'_0$ and never checked for $y'= y'_0$. 
    We obtain that
    \begin{align}
        \rho_+ \left( \mathbf{x}, \mathbf{x}' \right) &= 1 - P \left( Y'=y_0 | X'=\mathbf{x}' \right).
    \end{align}
    
\end{itemize}

Checking to which distribution $\iota$ belongs to also corresponds to having either a more confident prediction for $\mathbf{x}$ or $\mathbf{x}'$. 
They thus can also be determined in terms the entropies of $\hat{\boldsymbol\pi}$ and $\hat{\boldsymbol\pi}'$.

Since $Z = \Delta \left( \hat{\boldsymbol\pi}  \left( \mathbf{x} \right) , Y\right) -  \Delta \left( \hat{\boldsymbol\pi} \left(  \mathbf{x}' \right) , Y' \right) $ is a continuous random variable, checking that $\rho_+> \rho_-$ amounts to checking that $\rho_+>\frac{1}{2}$.
We can thus now deduce an optimal UQ-C-index ranking rule
\begin{align}
    r^*\left( \mathbf{x} , \mathbf{x}' \right) &=
        \omega\mathds{1}_{ p_{Y|X=\mathbf{x}} \left( y_0 \right)>\frac{1}{2} } + \left( 1-\omega \right) \mathds{1}_{ 1-p_{Y'|X'=\mathbf{x}'} \left( y_0 \right)>\frac{1}{2} },
        \\
    &=\omega \mathds{1}_{ \hat{y}\neq \hat{y}_{\textrm{bay}} } + \left( 1 - \omega \right) \mathds{1}_{ \hat{y}' = \hat{y}'_{\textrm{bay}} },
\end{align}

where $\omega = \mathds{1}_{ H \left( \hat{\boldsymbol\pi} \right) <  H \left( \hat{\boldsymbol\pi}' \right) } \in \left\{ 0;1 \right\}$ is a binary variable representing which input (from $\mathbf{x}$ and $\mathbf{x}'$) has the highest predicted entropy.

In order to further simplify the above, let us consider the $\omega=1$ case and assume:
\begin{align}
    & \hat{y}\neq \hat{y}_{\textrm{bay}} \\ 
    \text{and} \quad & H \left( \hat{\boldsymbol\pi} \right) <  H \left( \hat{\boldsymbol\pi}' \right).
\end{align}
We now investigate the following sub-cases:
\begin{itemize}
    \item $\hat{y}' = \hat{y}'_{\textrm{bay}}$: In this case, $\Vert \hat{\boldsymbol\pi}' -  \boldsymbol\delta_{\hat{y}'_{\textrm{bay}}} \Vert_1 < 1$ and $\Vert \hat{\boldsymbol\pi} -  \boldsymbol\delta_{\hat{y}_{\textrm{bay}}} \Vert_1 > 1$ where $\boldsymbol\delta_y$ denotes the one-hot encoded vector corresponding to distribution $\delta_y$.
    \item $\hat{y}' \neq \hat{y}'_{\textrm{bay}}$: 
    As both predictions are in disagreement with the Bayes classifier, we know that $0.5$ is between $\hat{\pi}'_{\hat{y}_{\textrm{bay}}}$ (resp. $\hat{\pi}_{\hat{y}_{\textrm{bay}}}$) and  $\hat{y}_{\textrm{bay}}'$ (resp. $\hat{y}_{\textrm{bay}}$). 
    In addition, $H \left( \hat{\boldsymbol\pi} \right) <  H \left( \hat{\boldsymbol\pi}' \right) \Rightarrow |\hat{\pi}'_{\hat{y}'_{\textrm{bay}}} - 0.5 | < |\hat{\pi}_{\hat{y}_{\textrm{bay}}} - 0.5 |$. 
    Combining the two pieces of information, we have $0.5  - \hat{\pi}'_{\hat{y}'_{\textrm{bay}}} < 0.5 - \hat{\pi}_{\hat{y}_{\textrm{bay}}}$ which also means $\Vert \hat{\boldsymbol\pi}' -  \boldsymbol\delta_{\hat{y}'_{\textrm{bay}}} \Vert_1 < \Vert \hat{\boldsymbol\pi} -  \boldsymbol\delta_{\hat{y}_{\textrm{bay}}} \Vert_1 $.
\end{itemize}

Next, we consider the $\omega=0$ case and assume:
\begin{align}
    & \hat{y}' = \hat{y}'_{\textrm{bay}}\\ 
    \text{and} \quad & H \left( \hat{\boldsymbol\pi} \right) \geq  H \left( \hat{\boldsymbol\pi}' \right)
\end{align}
We investigate the following sub-cases:
\begin{itemize}
    \item $\hat{y} = \hat{y}_{\textrm{bay}}$: 
    As both predictions are correct, we know that $\hat{\pi}'_{\hat{y}_{\textrm{bay}}}$ (resp. $\hat{\pi}_{\hat{y}_{\textrm{bay}}}$) is between $0.5$ and $\hat{y}_{\textrm{bay}}'$ (resp. $\hat{y}_{\textrm{bay}}$). 
    In addition, $H \left( \hat{\boldsymbol\pi} \right) \geq  H \left( \hat{\boldsymbol\pi}' \right) \Rightarrow |\hat{\pi}'_{\hat{y}'_{\textrm{bay}}} - 0.5 | \geq |\hat{\pi}_{\hat{y}_{\textrm{bay}}} - 0.5 |$.     
    Combining the two pieces of information, we have $\hat{\pi}'_{\hat{y}'_{\textrm{bay}}} - 0.5 \geq  \hat{\pi}_{\hat{y}_{\textrm{bay}}} - 0.5$ which also means $\Vert \hat{\boldsymbol\pi}' -  \boldsymbol\delta_{\hat{y}'_{\textrm{bay}}} \Vert_1 \leq \Vert \hat{\boldsymbol\pi} -  \boldsymbol\delta_{\hat{y}_{\textrm{bay}}} \Vert_1 $.    
    \item $\hat{y} \neq \hat{y}_{\textrm{bay}}$: In this case, $\Vert \hat{\boldsymbol\pi} -  \boldsymbol\delta_{\hat{y}_{\textrm{bay}}} \Vert_1 > 1$ and $\Vert \hat{\boldsymbol\pi}' -  \boldsymbol\delta_{\hat{y}'_{\textrm{bay}}} \Vert_1 < 1$.
\end{itemize}

In all cases, we have $\Vert\hat{\boldsymbol\pi}' - \hat{y}_{\textrm{bay}}' \Vert_1 \leq \Vert\hat{\boldsymbol\pi} - \hat{y}_{\textrm{bay}}\Vert_1$.
We are thus provided with the following implication:

\begin{align}
    r^*\left( \mathbf{x} , \mathbf{x}' \right)=1 \Rightarrow \Vert\hat{\boldsymbol\pi}' - \boldsymbol\delta_{\hat{y}_{\textrm{bay}}'} \Vert \leq \Vert\hat{\boldsymbol\pi} - \boldsymbol\delta_{\hat{y}_{\textrm{bay}}}\Vert
\end{align}

To prove the contraposition, assume $\Vert\hat{\boldsymbol\pi}' - \boldsymbol\delta_{\hat{y}_{\textrm{bay}}'} \Vert \leq \Vert\hat{\boldsymbol\pi} - \boldsymbol\delta_{\hat{y}_{\textrm{bay}}}\Vert$

We investigate the following cases:
\begin{equation}
\begin{cases}
    \Vert \hat{\boldsymbol\pi} -  \boldsymbol\delta_{\hat{y}_{\textrm{bay}}} \Vert_1 < 1 \quad \text{and} \quad \Vert \hat{\boldsymbol\pi}' -  \boldsymbol\delta_{\hat{y}'_{\textrm{bay}}} \Vert_1 < 1 \Rightarrow \hat{y}' = \hat{y}'_{\textrm{bay}}\\
    \Vert \hat{\boldsymbol\pi} -  \boldsymbol\delta_{\hat{y}_{\textrm{bay}}} \Vert_1 \geq 1 \quad \text{and} \quad \Vert \hat{\boldsymbol\pi}' -  \boldsymbol\delta_{\hat{y}'_{\textrm{bay}}} \Vert_1 \geq 1 \Rightarrow \hat{y}\neq \hat{y}_{\textrm{bay}} \\
    \Vert \hat{\boldsymbol\pi} -  \boldsymbol\delta_{\hat{y}_{\textrm{bay}}} \Vert_1 \geq 1 \quad \text{and} \quad \Vert \hat{\boldsymbol\pi}' -  \boldsymbol\delta_{\hat{y}'_{\textrm{bay}}} \Vert_1 < 1 \Rightarrow \hat{y}\neq \hat{y}_{\textrm{bay}} \quad \text{and} \quad \hat{y}' = \hat{y}'_{\textrm{bay}} \\
\end{cases}
\end{equation}

Note that the last case, $\Vert \hat{\boldsymbol\pi} -  \boldsymbol\delta_{\hat{y}_{\textrm{bay}}} \Vert_1 < 1$ and $\Vert \hat{\boldsymbol\pi}' -  \boldsymbol\delta_{\hat{y}'_{\textrm{bay}}} \Vert_1 > 1$ is 
not compatible under the initial assumption.

\begin{itemize}
    \item For the first case, using the symmetry of $H$, we have that the closer the prediction to the Bayes prediction is, the lower is the entropy, hence $H \left( \hat{\boldsymbol\pi} \right) \geq  H \left( \hat{\boldsymbol\pi}' \right)$. 
    \item For the second case, the closer the prediction from the Bayes prediction is, the higher is the entropy, hence $H \left( \hat{\boldsymbol\pi} \right) <  H \left( \hat{\boldsymbol\pi}' \right)$. 
    \item For the last case, either we have $H \left( \hat{\boldsymbol\pi} \right) <  H \left( \hat{\boldsymbol\pi}' \right)$, and $\hat{y}\neq \hat{y}_{\textrm{bay}}$, or we have $H \left( \hat{\boldsymbol\pi} \right) \geq  H \left( \hat{\boldsymbol\pi}' \right)$ and $\hat{y}' = \hat{y}'_{\textrm{bay}}$. 
\end{itemize}

In all the cases, we have $\Vert\hat{\boldsymbol\pi}' - \boldsymbol\delta_{\hat{y}_{\textrm{bay}}'} \Vert \leq \Vert\hat{\boldsymbol\pi} - \boldsymbol\delta_{\hat{y}_{\textrm{bay}}}\Vert \Rightarrow r^*\left( \mathbf{x} , \mathbf{x}' \right)=1  $.

We can conclude that an optimal ranking rule for the UQ-C-index is the one that checks if the prediction is closer to the Bayes prediction or not.

\begin{equation}
r^*(\mathbf{x}, \mathbf{x}') = 1 \Leftrightarrow  \frac{1}{2}\Vert \hat{\boldsymbol\pi}(\mathbf{x}) - \boldsymbol\delta_{\hat{y}_{\textrm{bay}} \left( \mathbf{x} \right)  } \Vert_1 \leq \frac{1}{2}\Vert  \hat{\boldsymbol\pi}(\mathbf{x}') - \boldsymbol\delta_{\hat{y}_{\textrm{bay}} \left( \mathbf{x}' \right)  }\Vert_1 
\end{equation}
And an associated scoring function is
\begin{equation}
s^*(\mathbf{x}) = \varphi \left( \mathbf{x} \right) = \frac{1}{2} \Vert \hat{\boldsymbol\pi}(\mathbf{x}) - \boldsymbol\delta_{\hat{y}_{\textrm{bay}} \left( \mathbf{x} \right)  }\Vert_1. 
\end{equation}

This proves that $\varphi $ is a maximizer of the UQ-C-index.
Using the same arguments as in the proof of \cref{thm:uq_auc} (see \cref{ap:proof_thm_uq_auc}), we can directly conclude that $s^*$ is a maximizer of UQ-C-index iff $s^* \in \mathcal{E}_{\varphi}$.

\section{PROOF OF \cref{thm:proof_nested_sets_c_index}}
\label{ap:proof_nested_sets_c_index}

In this appendix, one aims at showing that a scoring $s$ for which \eqref{eq:nested_sets_varphi} holds is not uniformly better than $s^* \in \mathcal{E}_s$. 
For the sake concision, we define $\Delta=\Delta\left( \hat{\boldsymbol\pi} \left( X \right), Y \right) $, $\Delta'=\Delta \left( \hat{\boldsymbol\pi} \left( X'\right), Y' \right) $, $S = s \left( X \right) $ and $S' = s \left( X' \right) $. 
Similarly as in \cref{ap:proof_phi_optim}, we denote $\tilde{s} = \psi_s^{-1} \circ s$ where $\psi_s$ is the strictly increasing function involved in \eqref{eq:nested_sets_varphi} and one than has $P \left(   \hat{Y}_{\textrm{bay}}\neq \hat{Y} | X \in \mathcal{L}_{\gamma} \left( \tilde{s} \right) \right) = \gamma, \forall \gamma \in  \mathscr{C}_{\text{mbc}} \left( \mathcal{E}_\varphi \right) $. 
Likewise, we define $\tilde{s}^*$ from $s^*$ in the same way. 

The UQ-C-index achieved by some scoring $s$ then writes
\begin{align}
    \textrm{UQ-C-index}\left(s \right) &= P \left( S \leq  S' | \Delta  < \Delta'    \right),\\
    &= P \left( \tilde{S} \leq  \tilde{S}' | \Delta  < \Delta'    \right),\\
    &= \int P \left( \tilde{S} \leq  \tilde{S}', \tilde{S}'=\gamma  | \Delta  < \Delta'\right) d\gamma,\\
    &= \int P \left( \tilde{S} \leq  \gamma | \Delta  < \Delta' \right) d\gamma,\\
    &= \int\int P \left( \tilde{S} \leq  \gamma, \Delta'=\alpha | \Delta  < \Delta' \right) d\gamma d\alpha,\\
    &= \int\int P \left( \tilde{S} \leq  \gamma | \Delta  < \alpha \right)P \left(\Delta'=\alpha | \Delta  < \Delta' \right)  d\gamma d\alpha,\\
    &= \int\int P \left( \tilde{S} \leq  \gamma | \Delta  < \alpha \right)\frac{P \left(\Delta'=\alpha , \Delta  < \Delta' \right)}{P \left( \Delta  < \Delta' \right) }  d\gamma d\alpha,\\
    &= 2 \int\int P \left( \tilde{S} \leq  \gamma | \Delta  < \alpha \right) P \left(\Delta  <  \alpha \right)  d\gamma d\alpha,\\
    &= 2 \int\int P \left( X \in  \mathcal{L}_\gamma \left( \tilde{s} \right)  | \Delta  < \alpha \right) P \left(\Delta  <  \alpha \right)  d\gamma d\alpha,\\
    &= 2 \int \underbrace{\int P \left( X \in  \mathcal{L}_\gamma \left( \tilde{s} \right)  | \Delta  < \alpha \right)   d\gamma}_{J \left( \alpha \right) } P \left(\Delta  <  \alpha \right)  d\alpha,    
    \label{eq:uq-c-index-interm}
\end{align}

Define $J^* \left( \alpha \right) = \int P \left( X \in  \mathcal{L}_\gamma \left( \tilde{s}^* \right)  | \Delta  < \alpha \right) d\gamma$. 
We have $J^* \left( \alpha \right) \leq J \left( \alpha \right)$ for any $\alpha$ as $\mathcal{L}_\gamma \left( \tilde{s}^* \right)  \subset \mathcal{L}_\gamma \left( \tilde{s} \right) $. 
In the remainder of this proof, $s^*$ and all other $*$-marked notations refer to this particular UQ-C-index maximizer. 

Let us also remark that function $J$ can be re-written as follows
\begin{align}
    J \left( \alpha \right) &= \int \frac{P \left( \tilde{S} \leq \gamma, \Delta < \alpha \right) }{P \left( \tilde{S} \leq \infty, \Delta < \alpha \right)}d\gamma.
\end{align}

We can check that the cdf of distribution $p_{\tilde{S},\Delta}$ is the main building brick in the definition of function $J$. 
Since $p_{S,\Delta}$ has no atom its cdf is continuous and so are the cdf of $p_{\tilde{S},\Delta}$ and function $J$. 
Likewise, since $p_{S^*,\Delta}$ has no atom, $J^*$ is continuous as well. 
Now we see that if there is an $\alpha_0$ for which $J^* \left( \alpha_0 \right) < J \left( \alpha_0 \right) $ then the UQ-C-index of $s$ would be larger than the UQ-C-index of $s^*$ as the strict inequality holds at least for a small neighborhood (of positive Lebesgue measure) around $\alpha_0$. 
The next paragraphs will show that $J^* \left( \frac{1}{2} \right) < J \left( \frac{1}{2} \right) $.

We have
\begin{align}
    J \left( \alpha \right) &= \int_0^1 \int_{\mathcal{L}_\gamma \left( \tilde{s} \right) } p_{X|\Delta < \alpha} \left( \mathbf{x} \right) d\mathbf{x} d\gamma, \\
    &= \frac{1}{P \left( \Delta < \alpha \right) } \int_0^1 \int_{\mathcal{L}_\gamma \left( \tilde{s} \right) } P \left( \Delta < \alpha | X=\mathbf{x} \right) p_X \left( \mathbf{x} \right) d\mathbf{x} d\gamma.
\end{align}

We also have 
\begin{align}
    \Delta \left(\hat{\boldsymbol\pi} \left( \mathbf{x} \right)  , Y \right) < \frac{1}{2} &\Leftrightarrow  1 - \hat{\pi}_Y  < \frac{1}{2},\\
     &\Leftrightarrow \hat{\pi}_Y > \frac{1}{2},\\
     &\Leftrightarrow Y=\hat{y}.
\end{align}
This implies that
\begin{align}
    J \left( \frac{1}{2} \right) &= \frac{1}{P \left( \Delta < \frac{1}{2} \right) } \int P \left( Y=\hat{Y}, X \in \mathcal{L}_\gamma \left( \tilde{s} \right)  \right) d\gamma.
\end{align}

Because $P \left( \Delta < \frac{1}{2} \right) = P \left( Y=\hat{Y} \right)$, we obtain
\begin{align}
    J \left( \frac{1}{2} \right) &= \int P \left( X \in \mathcal{L}_\gamma \left( \tilde{s} \right)  | Y=\hat{Y} \right) d\gamma,\\
      &>  \int P \left( X \in \mathcal{L}_\gamma \left( \tilde{s}^* \right)  | Y=\hat{Y} \right) d\gamma,\\
    &> J^* \left( \frac{1}{2} \right).
\end{align}

The penultimate inequality is obtained by leveraging the hypothesis that $s^* \prec_{\text{mbc}} s \Leftrightarrow \tilde{s}^* \prec_{\text{mbc}} \tilde{s}$ which involves the existence of a region $\mathscr{R}_{\tilde{s}^*\tilde{s}}$ with positive Lebesgue measure where $P \left( X \in \mathcal{L}_\gamma \left( \tilde{s} \right)  | \hat{Y}_{\textrm{bay}}=\hat{Y} \right) > P \left( X \in \mathcal{L}_\gamma \left( \tilde{s}^* \right)  | \hat{Y}_{\textrm{bay}}=\hat{Y} \right)$.  
We now can conclude that, under this hypothesis, $\tilde{s}$ would achieve a larger UQ-C-index than one maximizer of the latter which is not possible.

\section{PROOF OF \cref{lem:aurc}}
Similarly as in \cref{ap:proof_uq_cindex_optimal_score}, for $\left( X,Y \right)$ and $\left( X,Y \right) $ two random variable pairs i.i.d with respect to the data distribution $p_{X,Y}$, we denote $V = \mathds{1}_{ Y \neq \hat{Y} | X }$, $S=s \left( X \right) $ and $S' = s \left( X' \right) $. Starting from the definition of $\textrm{G-AUC}$, and for any fixed scoring function $s$, one can write:
\begin{align}
\textrm{G-AUC} \left(s \right) &= P \left( V=1 | S \leq S' \right). 
\end{align}

Since for any i.i.d. couple of random variables, one has $P \left( S \leq S' \right) = P \left( S' \leq S \right) = \frac{1}{2}$, we thus have 
\begin{align}
\textrm{AURC} \left(s \right) &= 2P \left( V=1 , S \leq S' \right),\\
&= 2P \left( V=1 , V'=0, S \leq S' \right) + 2P \left( V=1 , V'=1, S \leq S' \right).\label{eq:ap:aurc0}
\end{align}

We can remark that 
\begin{align}
    P \left( V=1 , V'=1, S \leq S' \right) &= P \left( S \leq S' | V = 1, V'=1 \right) P \left( V=1, V'=1 \right).  
\end{align}
Similarly as stated above, since $ \left( S|V=1 \right) $ and $\left( S'|V'=1 \right) $ are i.i.d, then $P \left( S \leq S' | V = 1, V'=1 \right) = P \left( S' \leq S | V = 1, V'=1 \right) = \frac{1}{2}$. 
In addition, we also have independence of $V$ and $V'$ and by definition of accuracy (for a fixed predictor $\hat{y}$), we thus obtain
\begin{align}
    P \left( V=1 , V'=1, S \leq S' \right) &= \frac{1}{2} \left( 1 - \textrm{ACC}  \right)^2. \label{eq:ap:aurc1}
\end{align}

Besides, one can also remark that 
\begin{align}
    P \left( V=1 , V'=0, S \leq S' \right) &= P \left( V' < V |  S \leq S' \right), \\
    &= P \left(  S \leq S' | V' < V \right) P \left( V' < V \right),\\
    &= \left( 1 - \underbrace{P \left( S' < S | V' < V \right)}_{=1 - \textrm{UQ-AUC}\left( s \right)  } \right) P \left( V' < V \right)\label{eq:ap:aurc2}
\end{align}

Moreover, we have 
\begin{align}
    P \left( V' < V \right) &= P \left( V' < 1 | V =1 \right)P \left( V=1 \right) + \underbrace{P \left( V' < 0 | V =0 \right)}_{=0} P \left( V=0 \right), \\
    &= P \left( V' = 0 | V =1 \right)P \left( V=1 \right), \\
    &= P \left( V' = 0\right)P \left( V=1 \right), \\
    &= \textrm{ACC}  \left( 1 - \textrm{ACC}   \right).\label{eq:ap:aurc3}
\end{align}

Finally, by substituting \eqref{eq:ap:aurc3} in \eqref{eq:ap:aurc2} and then \eqref{eq:ap:aurc2} and \eqref{eq:ap:aurc1} in \eqref{eq:ap:aurc0}, one obtains \eqref{eq:aurc}.

Concerning $\textrm{H-AUC}$, by definition, one has
\begin{align}
    \textrm{G-AUC} &= \int_0^{1 - \textrm{ACC}} P \left( S \leq F_{S|V=1}^{-1}\left( \frac{t}{1 - \textrm{ACC}} \right)  , V=0\right) dt,
\end{align}
where $F_{S|V=1}^{-1}$ is the inverse function of the cdf of $S|V=1$. 
Let us use the following change of variable under the integral:
\begin{align}
    u &:= F_{S|V=1}^{-1}\left( \frac{t}{1 - \textrm{ACC}} \right), \\
    \Leftrightarrow t &= F_{S|V=1}\left( u \right) \left( 1- \textrm{ACC} \right),\\
    \textrm{and } dt &= p_{S|V=1}\left( u \right)\left( 1- \textrm{ACC} \right) du.
\end{align}

We therefore obtain
\begin{align}
    \textrm{G-AUC} &= \int_{s_{\textrm{min}}}^{s_{\textrm{max}}} P \left( S \leq u  , V=0\right) p_{S|V=1}\left( u \right) \left( 1- \textrm{ACC} \right) du,\\
    &= \left( 1- \textrm{ACC} \right) \textrm{ACC} \int_{s_{\textrm{min}}}^{s_{\textrm{max}}} F_{S|V=0}\left( u \right) p_{S|V=1}\left( u \right) du,\\
\end{align}
where $s_{\textrm{min}}$ and $s_{\textrm{max}}$ are respectively the minimal and maximal values taken by $S$. 
Finally, since $\textrm{UQ-AUC}$ is also defined as $\int_{s_{\textrm{min}}}^{s_{\textrm{max}}} F_{S|V=0}\left( u \right) p_{S|V=1}\left( u \right) du$, one obtains \eqref{eq:h-auc}.

\section{ADJUSTED METRICS FOR OOD DETECTION}
\label{ap:ood}
In this appendix, we examine to what extent some of the results of this paper for the use case of misclassification error prone input detection can be transposed to the OOD input detection use case. 

We start with a situation in which a set of OOD data points is at hand. 
Since one can hold out a fraction of in-distribution (InD) points for evaluation purposes, it is thus possible to build a test dataset for which the label InD/OOD is available. 
Based on these labels, one can easily build an OOD analogue of ground truth $\phi$ and of UQ-AUC as done in prior arts. 
They will enjoy similar theoretical properties as those of \cref{thm:proof_nested_sets_phi}, \cref{thm:uq_auc} and \cref{thm:phi_optim}. 

The definition of an OOD analogue of the UQ-C-index appears to be less obvious (if not ill-advised) as one needs to define a meaningful gap function $\Delta \left( . \right)$ based on the network output $\hat{\boldsymbol\pi}\left( \mathbf{x} \right) $. 
One cannot directly compare softmax vectors with one-hot-encoding of the InD/OOD label. Consequently, it is necessary to first plug softmax vector into an OOD sensitive scoring function such as the free energy but this has the disadvantage of making the evaluation dependent of this choice. 
Another possibility is to train a companion model to discriminate InD from OOD data. 
Here again, the evaluation becomes dependent on modeling choices and training procedure. 

In the more challenging situation where no OOD is available to carry out the evaluation of the UQ score, it can envisioned to rely on distance informed versions of some metrics. 
For example, a C-index could be defined based the distance $\Delta \left( \mathbf{x} \right) = \underset{\mathbf{x}'\in \mathcal{I}_{\textrm{test}} \setminus \left\{ \mathbf{x} \right\}}{\min} d \left( \mathbf{x}, \mathbf{x}' \right)$ where $d$ is a distance on the input space and $\mathcal{I}_{\textrm{test}}$ is the set of inputs from the test set. 
For this approach to catch meaningful signals, the distance $d$ should intuitively rely on the data geometry otherwise the curse of dimensionality will come into play. 
Moreover, such a C-index will not directly inherit from the results of \cref{sub:c-index} as those rely on the definition of the gap function $\Delta$.

\section{EXPERIMENTAL DETAILS}
\label{app:exp-detail}
\subsection{Toy dataset}

Our experiments were carried out in around 10 hours of compute time on a 16-CPU and 256GB RAM machine where parallelization across CPUs was used. 
For the toy dataset, we sample scalars $\mu_0$ and $\mu_1$ from a 2-dimensional Gaussian distribution $\mathcal{N}(0, \tau I_2)$ where $I_2$ is the 2-by-2 identity matrix, with $\tau$ is a positive scalar controlling the overlap between $\mathcal{N}(\mu_1, \sigma)$ and $\mathcal{N}(\mu_0, \sigma)$. Then for each data point $i \leq n$, we sample the label $y^{(i)}$ from a Bernoulli distribution of parameter $p$. Then we sample the input feature from the corresponding Gaussian, $X^{(i)} \sim p_{X|Y=y^{(i)}} = \mathcal{N}(\mu_{y^{(i)}}, \sigma)$. 

This construction allows us to compute the exact input-conditional distribution:

\begin{equation}
    \boldsymbol\pi \left( \mathbf{x} \right)  = P(Y=y| X=x)  =
 \frac{p_{y}\: p_{X|Y=y}\left( \mathbf{x} \right)  }{\sum_{y' \in \mathcal{Y}} p_{y'} \: p_{X|Y=y'}\left( \mathbf{x} \right)}.
\end{equation}

\begin{figure*}[!h]
  \centering
  \begin{subfigure}[b]{0.32\textwidth}
    \includegraphics[width=\textwidth]{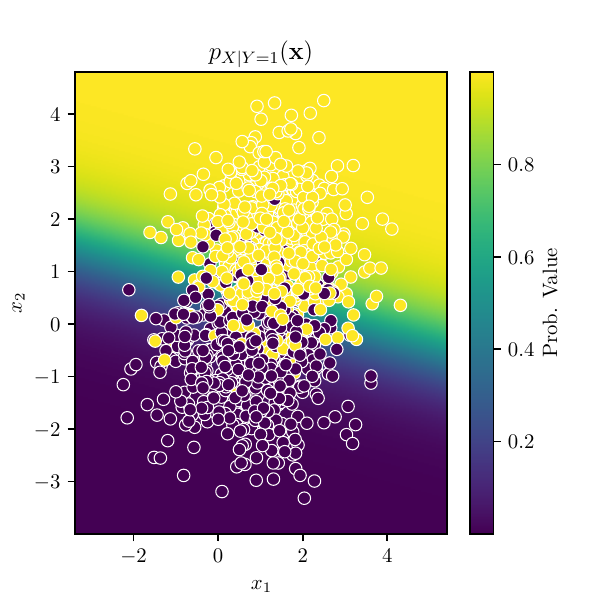}
    \caption{Data\label{fig:toy-dataset}}  
 \end{subfigure}
\begin{subfigure}[b]{0.32\textwidth}
    \includegraphics[width=\textwidth]{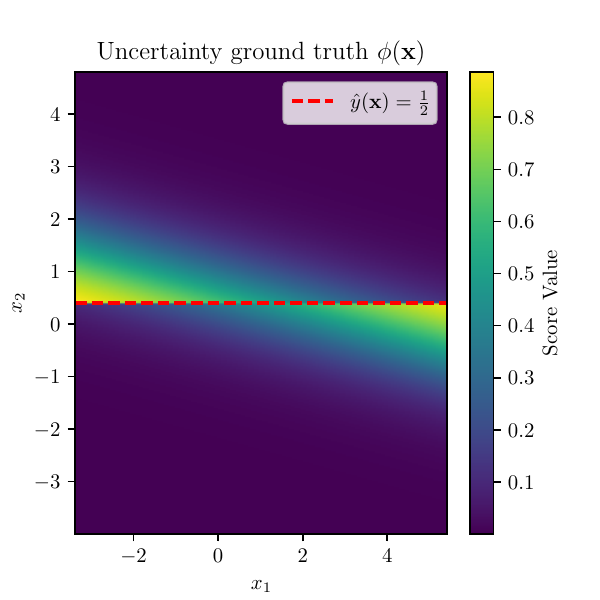}
     \caption{Example of function $\phi \left( \mathbf{x} \right) $\label{fig:phi_example}}
    \end{subfigure}
  \begin{subfigure}[b]{0.32\textwidth}
     \includegraphics[width=\textwidth]{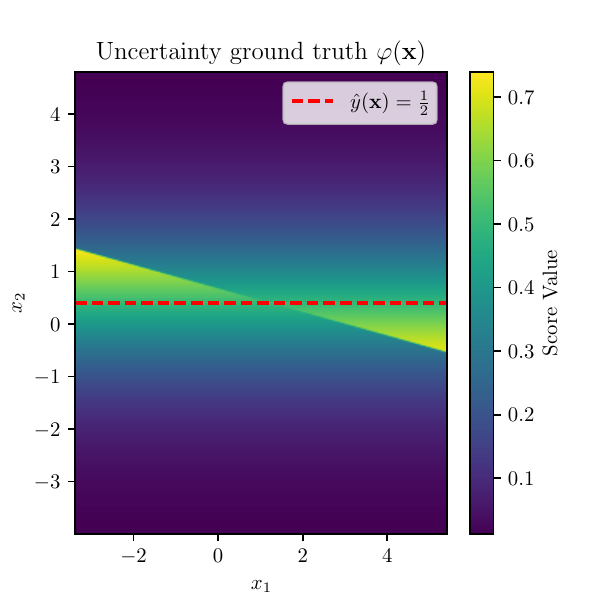}
     \caption{Example of function $\varphi \left( \mathbf{x} \right)$ \label{fig:varphi_example}} 
  \end{subfigure}
  \begin{subfigure}[b]{0.32\textwidth}
    \includegraphics[width=\textwidth]{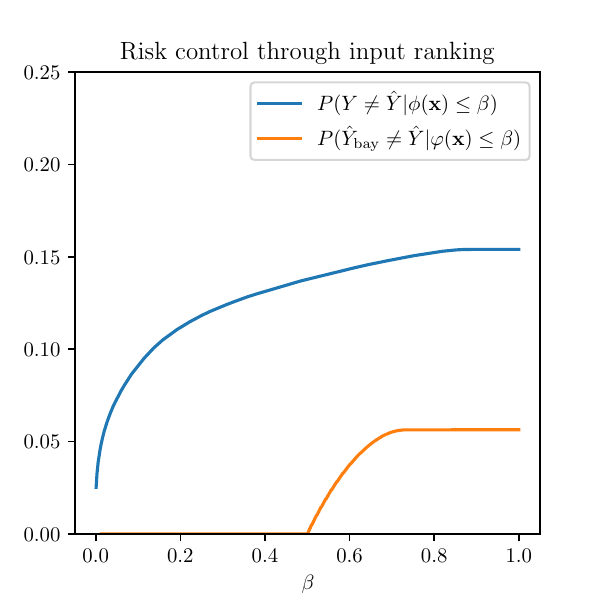}
    \caption{Risks incurred in sub-level sets\label{fig:risk_control}}
  \end{subfigure}

  \caption{Synthetic dataset illustrations. (a) Each dot represents a sample, whose color indicates the label. The background color indicates the probability $ P \left(Y= 1 | X=\mathbf{x} \right)$. (b) \& (c) Examples of ground truth scoring $\phi$ and $\varphi$ for an arbitrary predictor whose frontier decision is the dashed red line. (d) Different types of risk incurred in sub-level sets $\mathcal{L}_{\beta}\left( \phi \right) $ and $\mathcal{L}_{\beta}\left( \varphi \right) $ respectively. As implied by \cref{thm:proof_nested_sets_phi,thm:proof_nested_sets_varphi} those risks can be controlled by lowering the value of $\beta$.}
\end{figure*}

In the above, $p_1 = p$ and $p_0 = 1-p$. 
Distribution $\boldsymbol\pi \left( \mathbf{x} \right) $ allows us to compute the ground truth $\phi \left( \mathbf{x} \right) $ and $\varphi \left( \mathbf{x} \right) $. We use $n=1000$, $\tau=\sigma=1$ and $p_0=p_1= 0.5$. We use 600 samples for training, and 400 for testing, stratified by label. 
The dataset is displayed in \cref{fig:toy-dataset} along with examples of ground truth functions computed for an arbitrary predictor in \cref{fig:phi_example,fig:varphi_example}. 
This predictor uses $x^{(i)}_2 - \frac{1}{n}\sum_{j=1}^{n} x^{(j)}_2  $ as logits and its decision frontier is thus a horizontal line crossing the middle of the training sample.
Finally, \cref{fig:risk_control} shows the compliance of those ground truths with \cref{thm:proof_nested_sets_phi,thm:proof_nested_sets_varphi}.

Once the dataset has been sampled, we train an algorithm by minimizing the cross entropy, using hyper parameters described below.

Each algorithm is trained for 50 epochs, with 504 samples per batch. We used Adam optimizer, with learning rate set to either 0.005, 0.025 or 0.05. 
The models are MLPs with one or two hidden layers, of size 64, 32 or 16. 
The tested combinations are $[64, 32], [32, 32], [32, 16]$ and $[64]$.
For deep ensemble we used 5 or 10 models, and for MC-Dropout we sample either 10, 50 or 100 predictions. 
We used a dropout rate set to either 0.1, 0.3 or 0.5.

Lastly, we vary the scoring function among those described in \cref{sub:scoring_functions}. For DEs and MCD we used the entropy, the aleatoric entropy, the free energy, the mutual information, and the variance of the predicted class. For the softmax baseline, we used only the entropy, the free energy and the variance of the prediction. 

\begin{figure}[h]
    \centering
    \begin{subfigure}[b]{\textwidth}
       \centering
       \includegraphics[width=\textwidth]{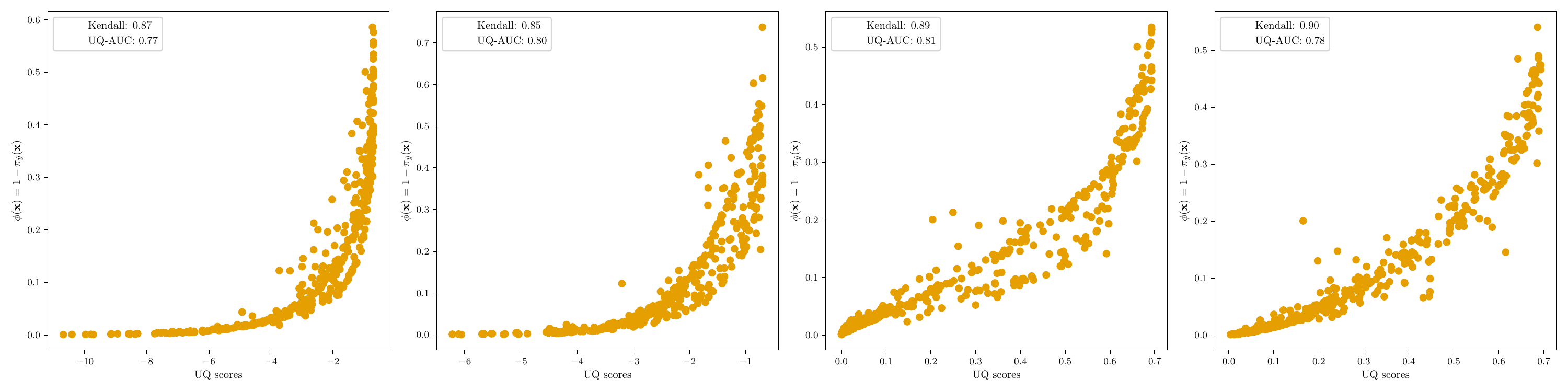}
       \caption{Softmax baselines}
       \label{fig:entropy-score-sft}
    \end{subfigure}

    \begin{subfigure}[b]{\textwidth}
       \includegraphics[width=\textwidth]{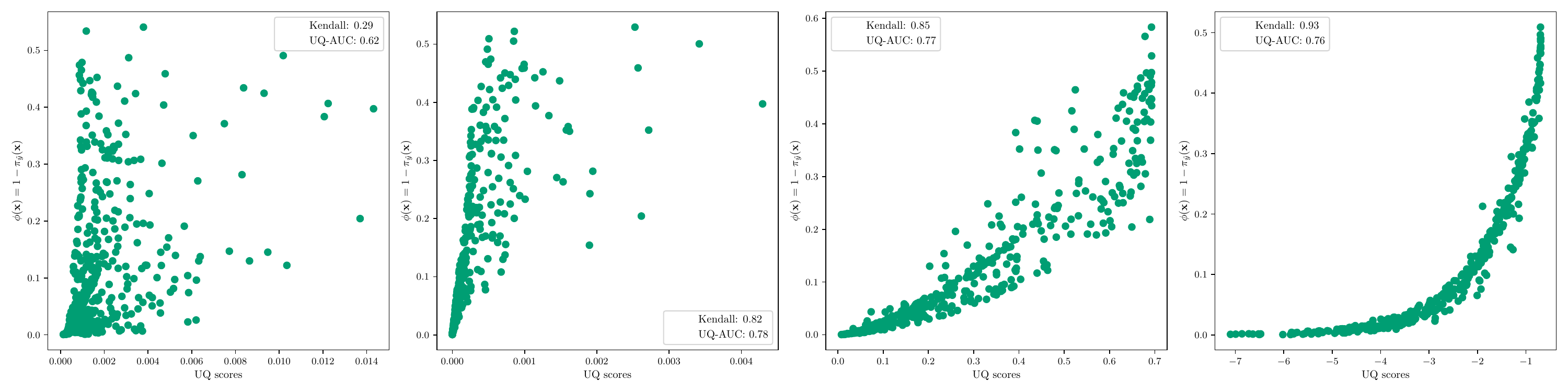}
             \caption{Deep ensembles}
       \label{fig:entropy-score-ens}
    \end{subfigure}

    \begin{subfigure}[b]{\textwidth}
       \includegraphics[width=\textwidth]{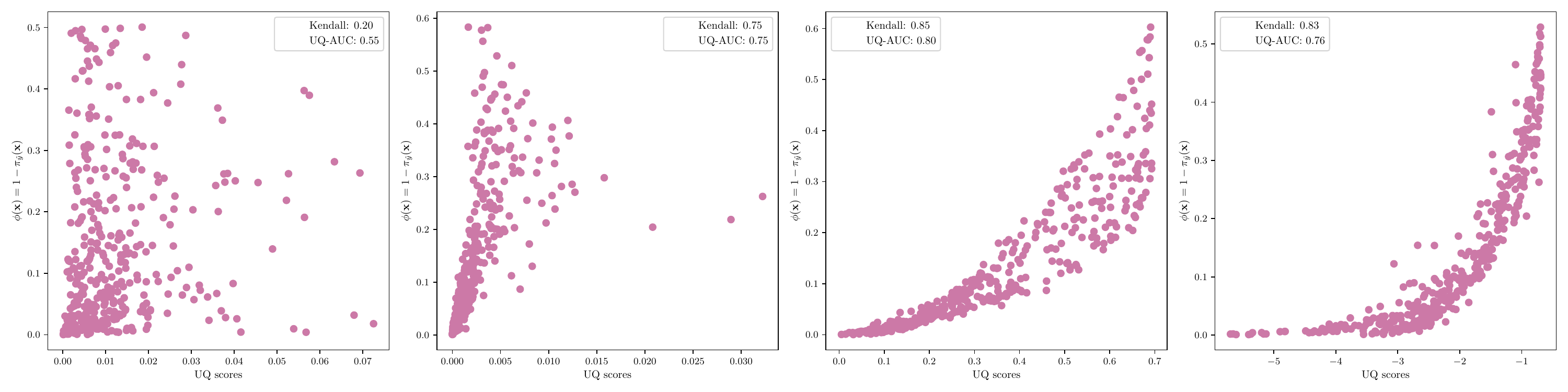}
             \caption{MC Dropout}
       \label{fig:entropy-score-mcd}
    \end{subfigure}

    \caption{Examples of scatter plots of UQ scores $s(\mathbf{x})$ versus ground truth $\phi(\mathbf{x})$ for randomly-picked trained UQ models. First row corresponds to the SFT baseline, second one are Deep Ensemble algorithms and the last one MC dropout algorithms. Each plot corresponds to one point in ~\cref{fig:cuq-misalign-bayes-label-synth}}
    \label{fig:phi-score}
 \end{figure}

\begin{figure}[h]
    \centering
    \begin{subfigure}[b]{\textwidth}
       \centering
       \includegraphics[width=\textwidth]{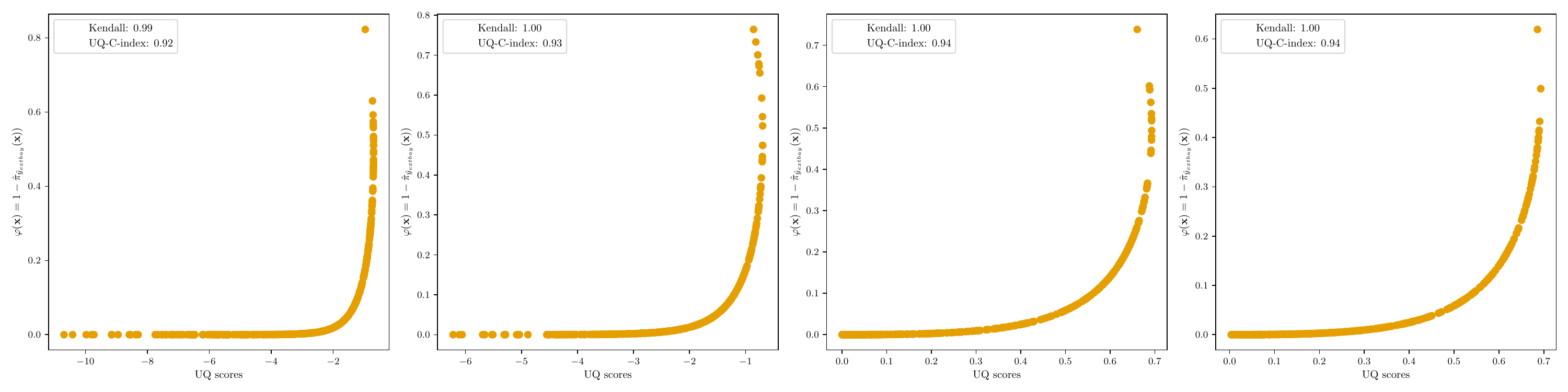}
       \caption{Softmax baselines}
       \label{fig:entropy-score-sft}
    \end{subfigure}

    \begin{subfigure}[b]{\textwidth}
       \includegraphics[width=\textwidth]{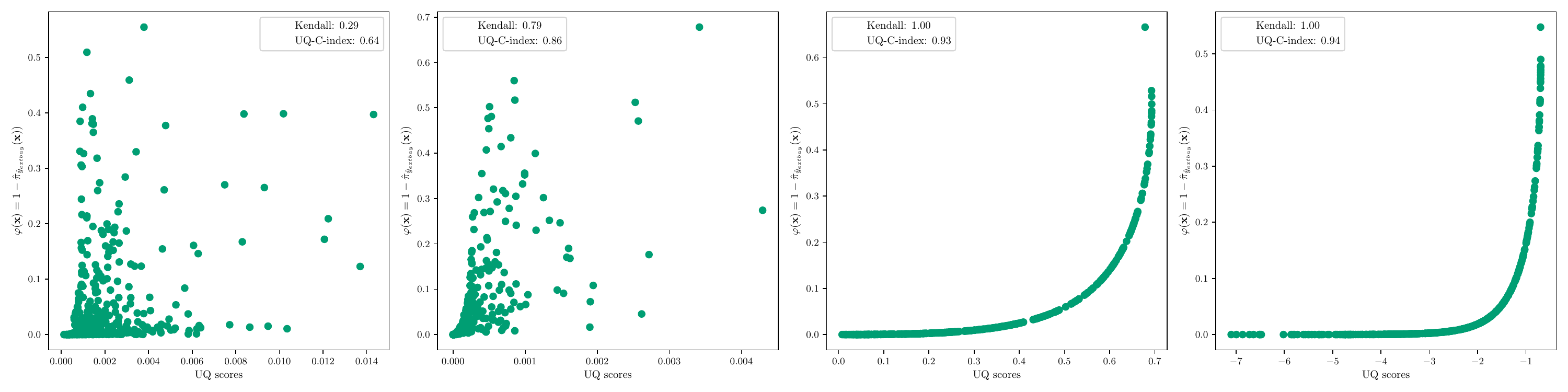}
             \caption{Deep ensembles}
       \label{fig:entropy-score-ens}
    \end{subfigure}

    \begin{subfigure}[b]{\textwidth}
       \includegraphics[width=\textwidth]{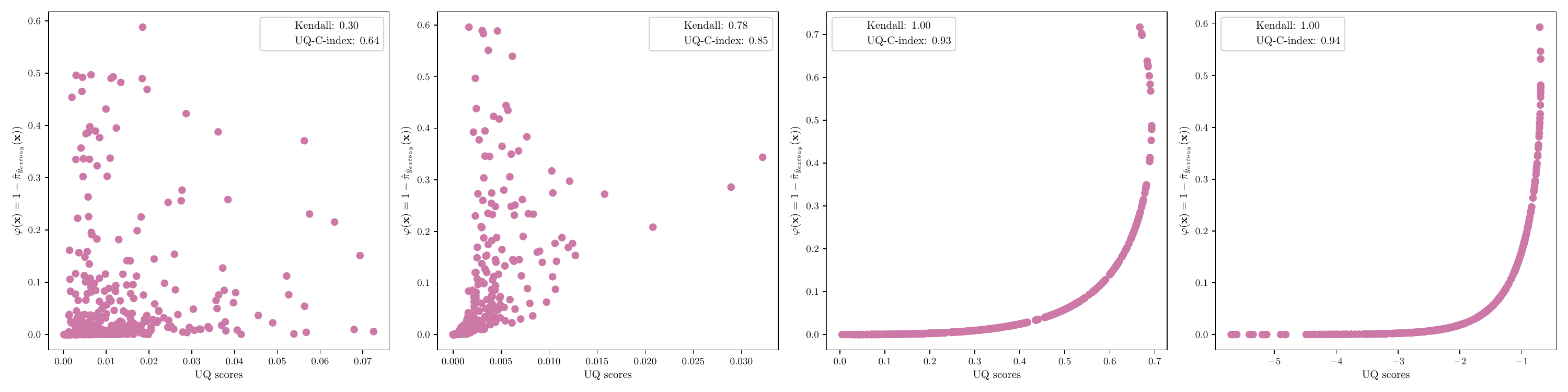}
             \caption{MC Dropout}
       \label{fig:entropy-score-mcd}
    \end{subfigure}

    \caption{Examples of scatter plots of UQ scores $s(\mathbf{x})$ versus ground truth $\varphi(\mathbf{x})$ for randomly-picked trained UQ models. First row corresponds to the SFT baseline, second one are Deep Ensemble algorithms and the last one MC dropout algorithms. Each plot corresponds to one point in ~\cref{fig:aucuq-bayes-error-synth}}
    \label{fig:varphi-score}
 \end{figure}

\begin{figure}[h]
    \centering
    \begin{subfigure}[b]{0.32\textwidth}
       \centering
       \includegraphics[width=\textwidth]{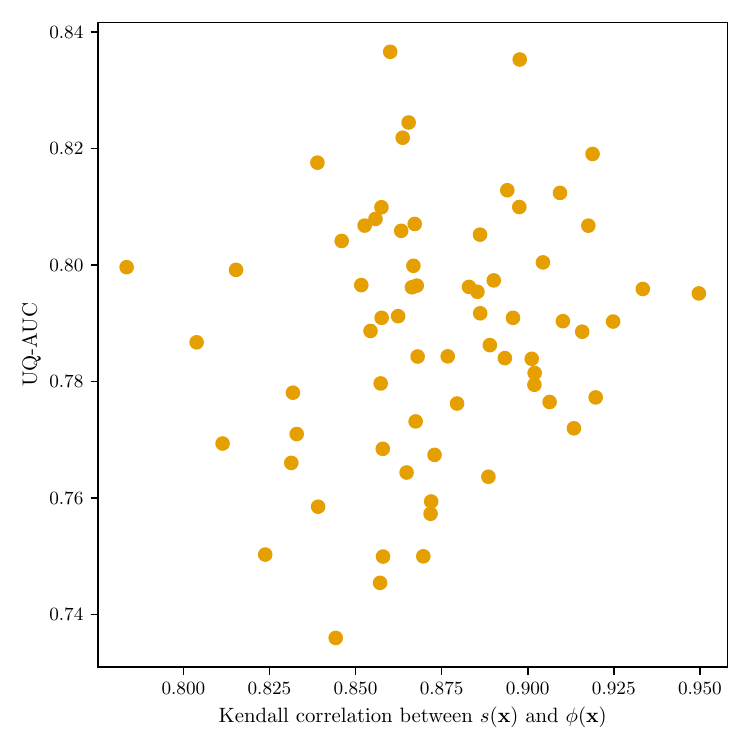}
       \caption{Softmax baselines}
       \label{fig:cindex-vs-speaman-ungrouped-sft}
    \end{subfigure}
    \begin{subfigure}[b]{0.32\textwidth}
       \includegraphics[width=\textwidth]{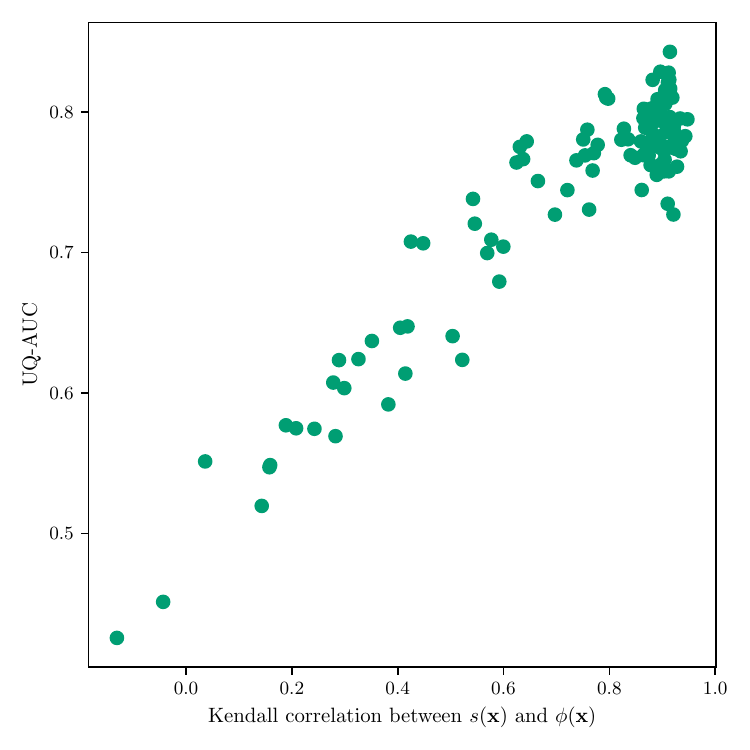}
             \caption{Deep ensembles}
       \label{fig:cindex-vs-speaman-ungrouped-ens}
    \end{subfigure}
    \begin{subfigure}[b]{0.32\textwidth}
       \includegraphics[width=\textwidth]{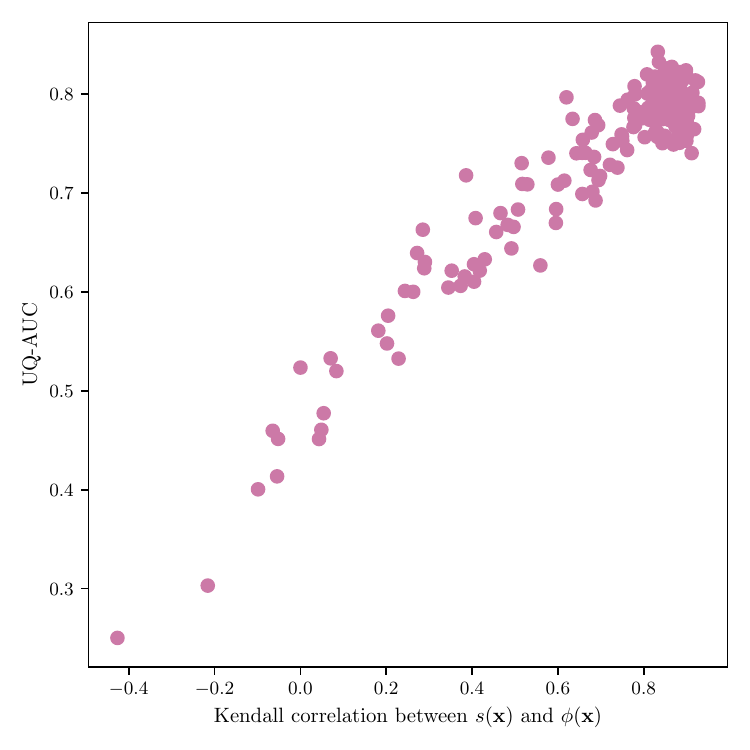}
             \caption{MC Dropout}
       \label{fig:cindex-vs-speaman-ungrouped-mcd}
    \end{subfigure}\\
    \begin{subfigure}[b]{0.32\textwidth}
       \centering
       \includegraphics[width=\textwidth]{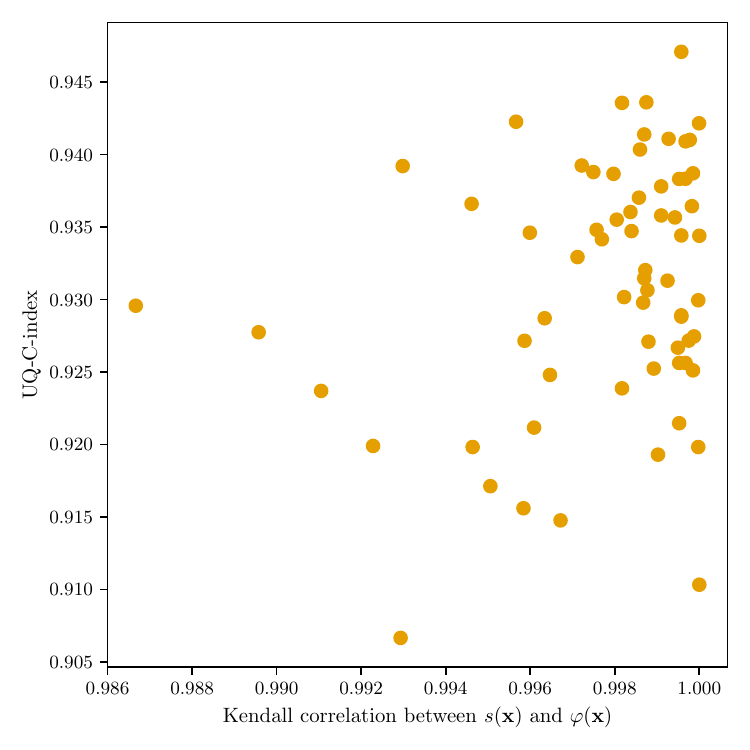}
       \caption{Softmax baselines}
       \label{fig:cindex-vs-speaman-ungrouped-sft}
    \end{subfigure}
    \begin{subfigure}[b]{0.32\textwidth}
       \includegraphics[width=\textwidth]{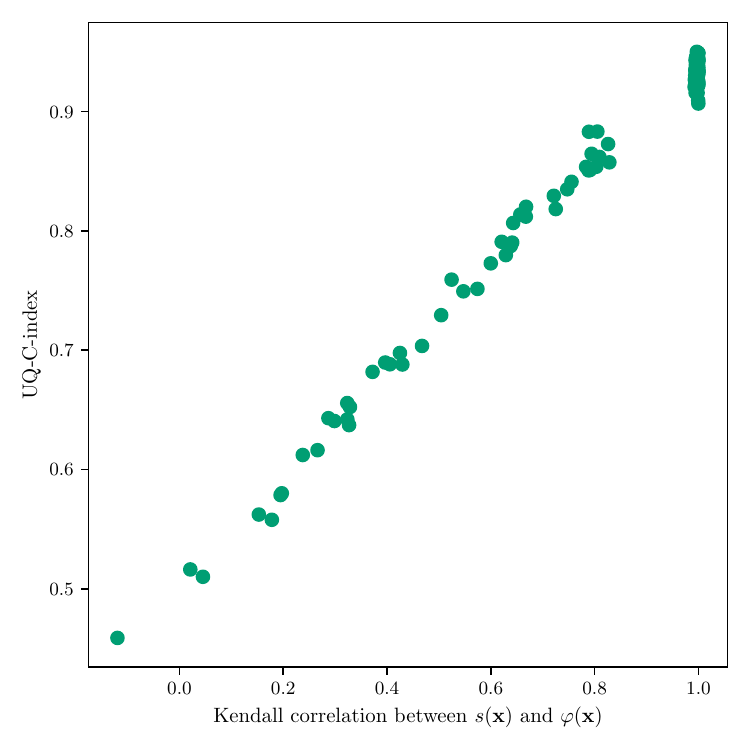}
             \caption{Deep ensembles}
       \label{fig:cindex-vs-speaman-ungrouped-ens}
    \end{subfigure}
    \begin{subfigure}[b]{0.32\textwidth}
       \includegraphics[width=\textwidth]{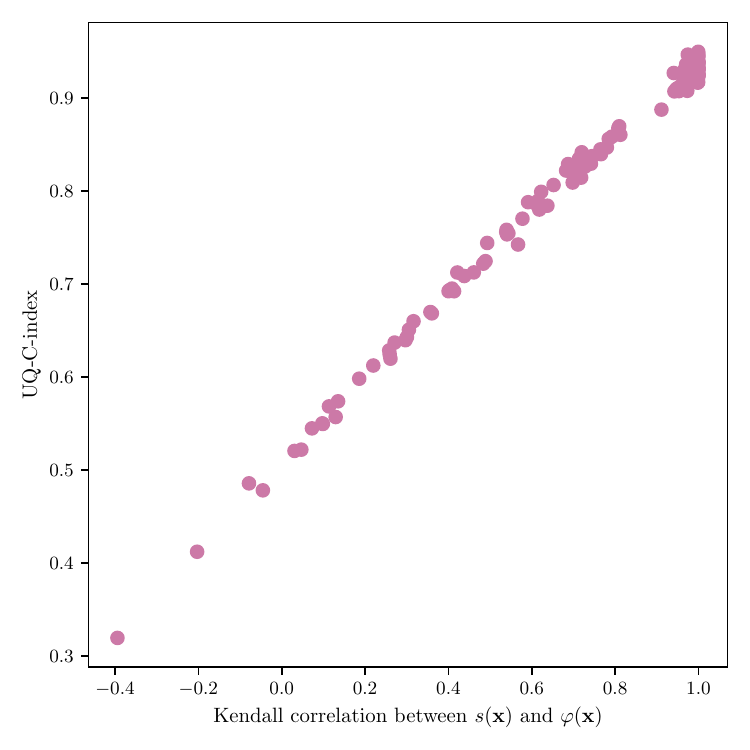}
             \caption{MC Dropout}
       \label{fig:cindex-vs-speaman-ungrouped-mcd}
    \end{subfigure}    

    \caption{UQ-AUC (resp. UQ-C-index) versus Kendall correlation between $s(\mathbf{x})$ and $\phi(\mathbf{x})$ (resp. $\varphi(\mathbf{x})$) for several UQ algorithms. All metrics are computed on a held-out test set. Note that Kendall correlations are intractable in practice, contrary to the UQ-AUC or UQ-C-index.}
    \label{fig:cindex-vs-speaman-ungrouped}
 \end{figure}

Once a model is trained, we compute the UQ score $s(\mathbf{x}^{(i)})$ and the ground truths $\phi \left( \mathbf{x}^{(i)} \right) $ and $\varphi \left( \mathbf{x}^{(i)} \right) $ for all samples in the test set, and then the Kendall correlation between those. 
\cref{fig:phi-score} and \cref{fig:varphi-score}  illustrate the distribution of the UQ scores versus each ground truth for several randomly-picked models. 
\cref{fig:cindex-vs-speaman-ungrouped} shows the same results as in \cref{fig:aucuq-bayes-error-synth,fig:cuq-misalign-bayes-label-synth} in different subplots (one for each UQ algorithm and UQ metric).

\subsection{Real-world datasets}
To scale our experiments to real-world datasets, we leverage two classical computer vision datasets with human annotations of uncertainty: CIFAR10-H~\citep{peterson2019human} and ReaL-ImageNet~\citep{beyer2020imagenet}.
We also leverage a large-scale dataset of text-to-text predictions with human judgments of uncertainty: OASST2~\citep{kopf2023openassistant}.
For CIFAR10-H, we use the same train/test split as in \citep{schmarje2022one}. The extra human annotations are obtained utilizing over 500k crowd-sourced human categorization judgments. We use the soft label distribution provided in~\citep{schmarje2022one}.
For the models, we leverage pre-trained ImageNet models available through the torchvision package. The complete list of models is provided in~\cref{subsection:list-of-models}.
To adapt these models to CIFAR10-H, we train linear probes on top of the penultimate layer of each network and report metrics computed on the validation folds. The training is conducted using SGD with a learning rate of 0.002 and momentum of 0.9 for 10 epochs with a batch size of 32. Probabilities $\hat{\boldsymbol\pi}\left( \mathbf{x} \right)$ are computed using the softmax function.

For ReaL-ImageNet, we use the pre-trained models without any fine-tuning and report metrics computed on the validation set from the original paper.
For the soft label distribution, we utilize the provided scores for each rated (image, label) pair, as computed by the \citet{dawid1979maximum} algorithm on the crowd-sourced votes.
We normalize these scores to sum to one across labels, yielding the soft label distribution. For more details on the data collection process and the computation of these scores, see \citep{beyer2020imagenet}.

The OASST2 dataset includes 6.6K text inputs for validation in various languages with several human annotations across 15 text content categories. Those categories were sorted into negative and positive labels to create a binary classification task. We performed inference using 70 pre-trained models available in the HuggingFace hub for zero-shot classification on the 15 categories. Likewise, model predictions were aggregated into softmax probabilities to address the binary classification task. For each model, we filtered out the sentences that were not in a language supported by the model.

As for the UQ scoring functions, we only consider the Entropy of the softmax output. 

Computation were done on a single T4 GPU for ImageNet as only inference is needed, taking approximately 1 hour.
For CIFAR10-H, we run each experiment on 15 T4 GPUs, taking approximately 15 hours.
For OASST2, we run our experiments parallelly on 8 V100 GPUs, taking approximately 10 hours. Thanks to zero-shot classification, no training is needed.

\subsubsection{List of models}%
\label{subsection:list-of-models}

Here are the list of models used in our experiments on CIFAR10-H and R-ImagetNet. 
We used the torchvision implementation of those models, with the available pretrained weights on ImageNet.

\begin{itemize}
    \item ResNet family: ResNet18, ResNet34, ResNet50, ResNet101, ResNet152
    \item ResNeXt family: ResNeXt50\_32x4d, ResNeXt101\_32x8d, ResNeXt101\_64x4d
    \item Wide ResNet family: Wide\_ResNet50\_2, Wide\_ResNet101\_2
    \item VGG family: VGG11, VGG11\_bn, VGG13, VGG13\_bn, VGG16, VGG16\_bn, VGG19, VGG19\_bn
    \item EfficientNet family: EfficientNet\_B0 to B7, EfficientNet\_V2\_S, EfficientNet\_V2\_M, EfficientNet\_V2\_L
    \item RegNet family: RegNet\_X (400MF, 800MF, 1.6GF, 3.2GF, 8GF, 16GF, 32GF), RegNet\_Y (400MF, 800MF, 1.6GF, 3.2GF, 8GF, 16GF, 32GF, 128GF)
    \item MobileNet family: MobileNet\_V2, MobileNet\_V3\_Small, MobileNet\_V3\_Large
    \item ShuffleNet family: ShuffleNet\_V2\_x0.5, ShuffleNet\_V2\_x1.0, ShuffleNet\_V2\_x1.5, ShuffleNet\_V2\_x2.0
    \item DenseNet family: DenseNet121, DenseNet161, DenseNet169, DenseNet201
    \item Vision Transformer (ViT) family: ViT\_B\_16, ViT\_B\_32, ViT\_L\_16, ViT\_L\_32, ViT\_H\_14
    \item Swin Transformer family: Swin\_T, Swin\_S, Swin\_B, Swin\_V2\_T, Swin\_V2\_S, Swin\_V2\_B
    \item ConvNeXt family: ConvNeXt\_Tiny, ConvNeXt\_Small, ConvNeXt\_Base, ConvNeXt\_Large
    \item MNASNet family: MNASNet0.5, MNASNet0.75, MNASNet1.0, MNASNet1.3
    \item Other models: AlexNet, GoogLeNet, Inception\_V3
\end{itemize}

As for the models used for OASST2, we used the following models:

\begin{itemize}
    \item BART family: BART-large-mnli, DistilBART-mnli-12-1, DistilBART-mnli-12-3, DistilBART-mnli-12-6, DistilBART-mnli-12-9
    \item BERT family: BERT-base-multilingual-cased, BERT-base-turkish-cased, BERT-base-spanish-wwm-cased, NB-BERT-base, MobileBERT-uncased
    \item RoBERTa family: RoBERTa-base, RoBERTa-large, DistilRoBERTa-base, XLM-RoBERTa-base, XLM-RoBERTa-large
    \item DeBERTa family: DeBERTa-v3-base, DeBERTa-v3-large, DeBERTa-v3-xsmall, mDeBERTa-v3-base
    \item DistilBERT family: DistilBERT-base-uncased, DistilBERT-base-multilingual-cased
    \item MiniLM family: MiniLM2-L6-H768, MiniLMv2-L6, MiniLMv2-L12
    \item E5 family: E5-base, E5-large, E5-large-instruct
    \item mContriever family: mContriever-msmarco
    \item Other models: FLAN-T5-base, CamemBERT-base, RuBERT-base, Megatron-BERT-large
\end{itemize}


\end{document}